\documentclass[journal]{IEEEtran}

\usepackage{amsmath,amssymb,amsfonts}
\usepackage{mathtools}
\usepackage{amsthm}
\newtheorem{proposition}{Proposition}
\usepackage{enumitem}
\usepackage{graphicx}

\ifCLASSOPTIONcompsoc
  \usepackage[caption=false,font=normalsize,
              labelfont=sf,textfont=sf]{subfig}
\else
  \usepackage[caption=false,font=footnotesize]{subfig}
\fi

\usepackage{booktabs}
\usepackage{multirow}
\usepackage{makecell}
\usepackage{array}
\usepackage{tabularx}
\usepackage{adjustbox}

\usepackage{pifont}
\usepackage{gensymb}
\usepackage[mathscr]{euscript}

\usepackage{algorithmic}

\usepackage{cite}

\usepackage[hidelinks]{hyperref}
\hypersetup{
     colorlinks = true,
     urlcolor = black
}

\usepackage[table,xcdraw,x11names,svgnames,dvipsnames]{xcolor}
\usepackage{color}

\usepackage{colortbl}

\ifCLASSINFOpdf
\else
\fi

\begin{document}
% \title{ASM-6D: Generative Correspondences for Fast Object Pose--Shape Fitting}
% \title{ASM-6D: Generative Complete Correspondences\\for Joint Pose–Shape Inference}

% \title{\nf{ASM-6D: Fast Certified Joint Pose–Shape Inference from Generative Complete Correspondences}}
\title{ASM-6D: Fast Certified Pose–Shape Inference with Complete Correspondence Flow Maps}


% \author{Anonymous Author(s)}% <-this % stops a space
\author{Ho Jin Choi, Minku Kim, and Nadia Figueroa%
\thanks{Ho Jin Choi and Minku Kim contributed
equally to this work.}%
\thanks{Ho Jin Choi and Nadia Figueroa are with
the University of Pennsylvania,
Philadelphia, PA, USA
(e-mail: cr139139@engineering.upenn.edu;
nadiafig@engineering.upenn.edu).}%
\thanks{Minku Kim is with Oregon State University,
Corvallis, OR, USA
(e-mail: kimminku@oregonstate.edu).}%
}

\maketitle

\begin{abstract}
% Partial RGB-D observations may admit multiple object shapes and poses. We introduce ASM-6D, which represents alternative interpretations as complete 3D correspondence sets with fixed indices shared with category-level Active Shape Models (ASMs). A conditional flow-based model samples the locations of visible and occluded template points, and each sample is fitted to an ASM to recover rotation, translation, isotropic metric scale, and shape coefficients. By analytically eliminating translation and shape, the fitting problem reduces to a nine-variable nonconvex scaled-rotation problem that can be solved efficiently in batch with ADMM. On ROPE, our four-step model achieves an IoU$_{50}$ AUC of 30.95 and a $10^\circ$/5 cm VUS of 39.14, compared with the reported RFMPose results of 21.0 and 30.8, respectively. Experiments on HouseCat6D and controlled ablations further evaluate pose accuracy, the contribution of active shape variation, and alternative shape--pose explanations under occlusion. The same representation also supports shape completion, geometry-based prior selection, temporal tracking, and qualitative grasping. Overall, ASM-6D provides a structured interface between learned geometric hypotheses and explicit pose--shape inference.

Category-level 6D pose and shape estimation requires recovering an object's pose and geometry from partial observations without an exact instance model. We present ASM-6D, which couples generative complete-geometry prediction with explicit category-level pose--shape inference through shared correspondences. Given an RGB-D image, ASM-6D generates alternative complete 3D correspondence hypotheses whose point identities are shared with a category-level Active Shape Model (ASM), including occluded points. Each hypothesis is fitted directly for rotation, translation, metric scale, and instance shape without a separate correspondence-assignment stage. We analytically eliminate translation and shape coefficients, reducing fitting to a nine-variable scaled-rotation optimization problem that enables efficient batched evaluation of multiple hypotheses in real time. ASM-6D achieves state-of-the-art performance on ROPE and HouseCat6D benchmarks. Ablations demonstrate the importance of generative correspondence prediction and active shape variation. The representation also supports shape completion, shape-dependent pose ambiguity, tracking, and robotic grasping at 6-38~Hz, depending on the inference budget. Project website: \url{https://asm6d-project.github.io/}

% Category-level 6D pose and shape estimation requires recovering an object’s pose and geometry from partial observations without access to its exact instance model. We present ASM-6D, which represents alternative interpretations of an RGB-D observation as complete indexed 3D correspondence sets. Each predicted point shares a fixed index with a category-level Active Shape Model, including points hidden by occlusion, allowing every generated hypothesis to be fitted directly for rotation, translation, metric scale, and instance shape. We analytically eliminate translation and shape coefficients, reducing joint fitting to a compact nine-variable scaled-rotation problem that enables efficient batched evaluation of multiple hypotheses. ASM-6D achieves state-of-the-art performance on both the ROPE and HouseCat6D benchmarks. Controlled ablations demonstrate the importance of generative correspondence prediction and active shape variation. The resulting indexed representation additionally supports shape completion, shape-dependent pose ambiguity, temporal tracking, and downstream robotic grasping.
\end{abstract}

\begin{IEEEkeywords}
Active shape models, category-level 6D object pose estimation, indexed correspondences, pose--shape optimization, RGB-D perception.
\end{IEEEkeywords}

% \fi
% \IEEEpeerreviewmaketitle

\section{Introduction}
\label{sec:intro}
Category-level 6D pose and shape estimation seeks to recover an object’s pose, metric scale, and geometry without access to its exact instance model. For robotic interaction, pose alone does not describe the surfaces available for manipulation, while RGB-D observations often reveal only part of the object. A useful representation must therefore recover complete geometry while retaining the structure needed for joint pose--shape inference, and represent alternative interpretations when occlusion, symmetry, or intra-category variation makes the observation ambiguous, all the while running in real time to be useful for downstream robotic manipulation tasks.

Existing approaches provide complementary pieces of this problem. Canonical-coordinate methods associate observed points with a normalized object frame \cite{wang2019normalized}, while completion-based methods recover missing geometry, including completed-keypoint representations \cite{Chen2020LearningCS,Irshad2022ShAPOIR,ren2026compose}. Generative pose methods instead retain alternative solutions directly in transformation space \cite{Zhang2024Omni6DPoseAB,ouyang2026rfmpose}. Active Shape Models (ASMs) provide a compact representation for estimating pose and instance shape when indexed correspondences are available \cite{Cootes1995ActiveSM,shi2023optimal,zhang2025shapeicp,shaikewitz2025category}. However, learned completion and explicit pose--shape inference remain largely separate. Completion methods recover geometry, while ASM-based inference assumes that correspondences are already available. This gap is particularly
relevant for object manipulation in dynamic environments, where inference must also remain computationally practical; recent generative category-level methods operate around $9$--$11$ Hz in reported tracking settings ~\cite{Zhang2024Omni6DPoseAB,ouyang2026rfmpose}.
% \nfcomment{Add a sentence about the current computation time of SOTA approaches related to the real-time constraints necessary for object manipulation in dynamic environments.}

\begin{figure}[!t]
    \centering
    \includegraphics[width=\linewidth]
    {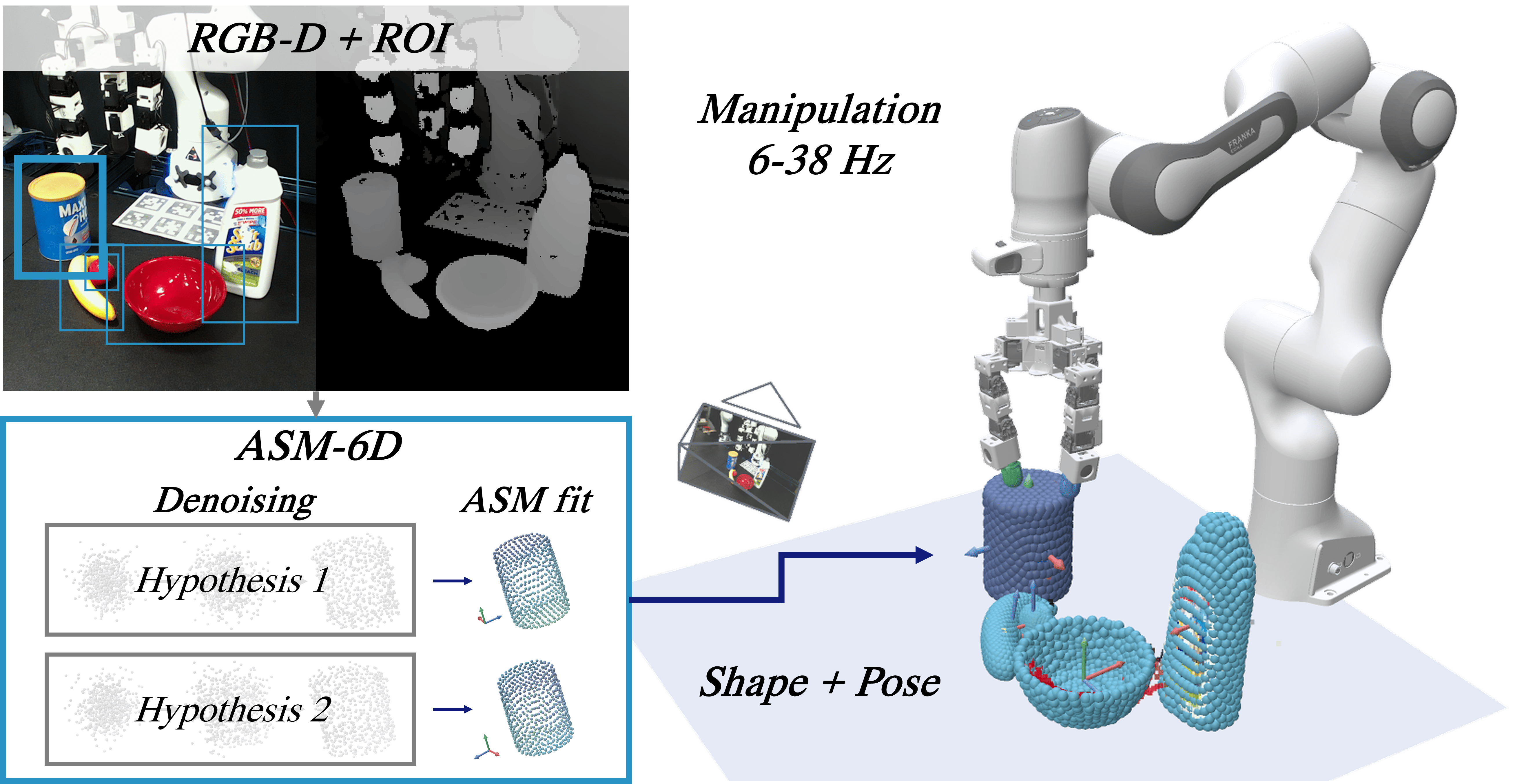}
    \caption{ASM-6D generates alternative complete indexed 3D correspondence hypotheses, whose shared point identities enable direct ASM fitting for pose, scale, and instance shape with a computation rate of 6--38~Hz. 
    % \nfcomment{Add Hz and multiple hypotheses in figure.}
    }
    \label{fig:intro}
    \vspace{-0.5em}
\end{figure}

ASM-6D bridges these reliability and computational gaps by learning a distribution over complete camera-frame correspondences whose indexing is shared with a category-level ASM. Each sample is both a complete geometric interpretation of the observation and a directly solvable pose--scale--shape hypothesis that can be computed in real-time (up to 38.85 Hz in the lowest-budget known-prior
setting; Table~\ref{tab:full_pipeline_fps}) while
taking an RGB-D ROI directly, without requiring a mask-segmented point cloud. ASM-6D represents each object interpretation as a complete indexed \emph{correspondence hypothesis} (Fig.~\ref{fig:intro}). A hypothesis predicts a camera-frame 3D location for every point of a shared template, including occluded points, with predicted point \(i\) corresponding directly to ASM point \(i\). The indexing is constructed offline by deforming a common template onto normalized training instances while preserving vertex identities; the same indices define the ASM and generator targets. Each hypothesis therefore provides complete candidate geometry together with the correspondences required for direct pose, metric-scale, and shape fitting. Multiple samples represent alternative interpretations of ambiguous observations.

Converting the generated correspondences into metric pose and shape requires jointly estimating rotation, translation, scale, and ASM deformation. For fixed scaled rotation, the objective is quadratic and strictly convex in translation and shape coefficients, allowing both to be eliminated in closed form, reducing scale-aware ASM fitting to a nine-variable scaled-rotation problem.
% \nfcomment{Based on what assumption or insight can you reach this analytical elimination, explain in a short phrase.}
A batched ADMM (Alternating Direction Method of Multipliers) solver~\cite{Boyd2011DistributedOA} evaluates correspondence hypotheses and, when needed, candidate shape priors in parallel, after which a visibility-weighted geometric residual selects the final estimate, in real time. 

In addition to offering reliability and computational efficiency, ASM-6D achieves state-of-the-art category-level 6D pose estimation performance on both ROPE \cite{Zhang2024Omni6DPoseAB} and HouseCat6D \cite{jung2024housecat6d} under their respective evaluation protocols. On ROPE, ASM-6D improves IoU$_{50}$ AUC from 21.0 to 30.95 over RFMPose~\cite{ouyang2026rfmpose}. In controlled ablation experiments, we show benefits from generative correspondence prediction and active shape variation. We additionally evaluate the capabilities of ASM-6D on shape completion, shape-conditioned pose ambiguity, prior selection, tracking, and robotic grasping.
% \nfcomment{Showing what improvement over SOTA? List some numbers here of relevant metrics from these experiments. Show off the performance more!}

\noindent Thus, our contributions are the following:
\begin{itemize}[leftmargin=*]
% \item \textbf{Generative indexed correspondence representation:} We introduce complete 3D correspondence hypotheses that predict both visible and occluded geometry while preserving the fixed point identities required for direct category-level ASM fitting.
% \item \textbf{Generative complete-geometry representation:}
% We generate alternative complete 3D hypotheses in a correspondence space shared with a category-level ASM, enabling direct joint estimation of pose, metric scale, and instance shape.
\item \textbf{Generative complete correspondences for pose--shape inference:}
We learn a distribution over complete camera-frame correspondences in the same indexed space as a category-level ASM, so each sample can be fitted directly for pose, metric scale, and instance shape. For training, we use the Omni6DPose~\cite{Zhang2024Omni6DPoseAB} SOPE dataset (475K images, 4,385 CAD models across 156 categories), followed by evaluation and adaptation on HouseCat6D~\cite{jung2024housecat6d}.
% \nfcomment{From which datasets? How big is the training data and how many object categories does it include?}

\item \textbf{Scale-aware joint pose--shape inference:} We analytically reduce joint pose, metric-scale, and ASM-shape estimation to a nine-variable scaled-rotation optimization problem and solve many hypothesis--prior fits efficiently in batch, enabling 6--38 Hz full-pipeline inference depending on the inference budget.
% \nfcomment{Highlight computation time here.}

\item \textbf{Empirical validation and representation analysis:} ASM-6D achieves state-of-the-art pose performance on ROPE~\cite{Zhang2024Omni6DPoseAB} (30.95\% IoU$_{50}$) and HouseCat6D~\cite{jung2024housecat6d} (85.7\% IoU$_{50}$), alongside 73.7\% exact 156-way prior selection on RealNOCS~\cite{wang2019normalized}. Controlled ablations and experiments demonstrate completion accuracy under occlusions, robust handling of rotational ambiguity, and real-time tracking and grasping experiments, characterizing the representation.
% \nfcomment{Re-write this bullet point providing the concrete performance metrics on all the experiments you list.}

\end{itemize}

\noindent \textbf{Paper Organization:}
Section~II reviews related work. Sections~III--V introduce the indexed representation, correspondence flow maps, and scale-aware fitting; Sections~VI--VII evaluate pose estimation and representation reuse, with implementation details and derivations in the appendices.

\section{Related Work}
\label{sec:related_works}

\noindent\textbf{Canonical correspondences and shape representations.} Several recent methods establish geometric structure from the observation. NOCS \cite{wang2019normalized} predicts canonical coordinates for observed object pixels, establishing correspondences between image observations and a normalized object frame. Subsequent approaches deform category-level shape priors \cite{tian2020shape} or learn canonical and implicit shape representations for joint pose, size, and geometry estimation \cite{Chen2020LearningCS,Irshad2022ShAPOIR}. While these methods improved shape estimation performance, their correspondence representations are primarily tied to observed image evidence or canonical object space. ASM-6D instead predicts the camera-frame locations of a complete, fixed-index template, including occluded points. The resulting geometry preserves the point identities required for direct joint pose--shape fitting.

\noindent\textbf{Completion-based pose estimation.} Point-cloud completion methods reconstruct missing geometry from partial observations \cite{yu2021pointr,xiang2021snowflakenet, yu2023adapointrdiversepointcloud}, but completion alone does not specify how the recovered geometry should enter explicit category-level pose--shape inference. ComPose \cite{ren2026compose} more closely couples the two tasks by progressively recovering keypoints and dense geometry together with canonical correspondences. ASM-6D instead represents each generated completion as a complete fixed-index correspondence set shared directly with a parametric category model, enabling explicit fitting for pose, metric scale, and instance shape without a separate correspondence-assignment stage.

\noindent\textbf{Generative pose estimation.} GenPose \cite{zhang2024generative} models category-level pose distributions using score-based diffusion, while RFMPose \cite{ouyang2026rfmpose} and Flow6D \cite{mei2026flow6d} use flow-based generation or refinement in transformation space.
% These methods retain alternative pose solutions under ambiguous observations, but the sampled variable is the object transformation itself.
Pose-space samples retain transformation ambiguity but do not explicitly represent the completed geometry or indexed point identities supporting each pose, limiting direct pose--shape fitting.
% \nfcomment{Explain why is this a problem or undesirable to estimate?}

ASM-6D instead generates complete indexed geometry, so each sample specifies both a possible pose after fitting and the complete instance geometry associated with that interpretation.

\noindent\textbf{Geometric pose--shape fitting.} Active Shape Models (ASMs) \cite{Cootes1995ActiveSM} represent intra-category variation using a low-dimensional linear shape basis. Shi et al. \cite{shi2023optimal} jointly estimate pose and shape from semantic keypoints using a certifiable semidefinite relaxation, while ShapeICP \cite{zhang2025shapeicp} jointly refines pose, shape, and correspondence. Recent specialized solvers further demonstrate efficient ASM fitting with a posteriori optimality certification \cite{shaikewitz2025category}. These methods address geometric inference once suitable observations or correspondences are available. ASM-6D complements this line by generating the complete indexed correspondences required by the shape model and jointly fitting pose, metric scale, and instance shape across multiple hypotheses and candidate priors.

In contrast to methods that address correspondence generation or pose--shape fitting separately, ASM-6D couples both through a shared indexed representation that remains practical for online inference.

\section{Problem and Indexed Representation}
\label{sec:problem_formulation}

\begin{figure*}[!t]
    \centering
    \includegraphics[width=\textwidth]{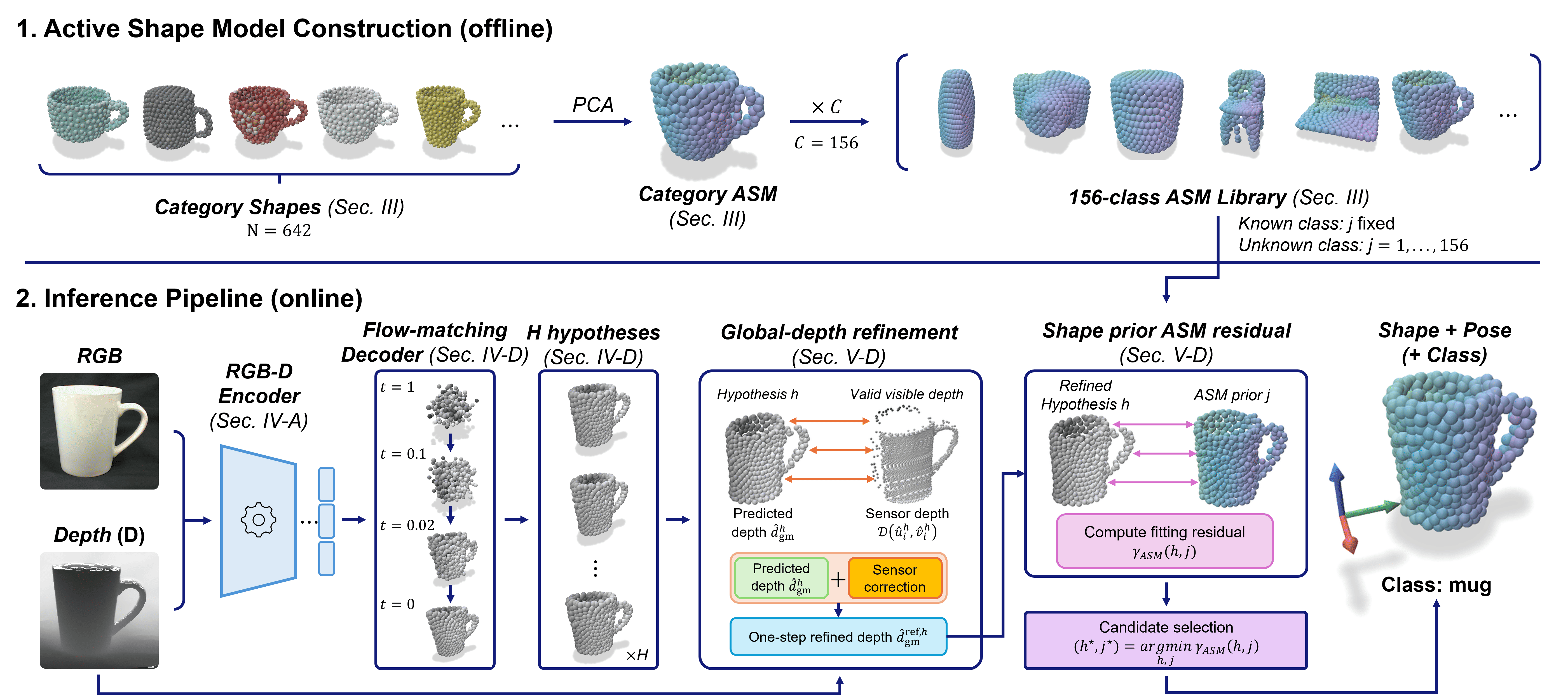}
    \caption{\textbf{Overview of ASM-6D.} Standardized training shapes define category-level Active Shape Models (Sec~\ref{sec:problem_formulation}). Given an RGB-D crop, the model generates multiple hypotheses of the complete indexed correspondence set (Sec~\ref{sec:complete_correspondence}). Each hypothesis is reconstructed in metric coordinates and fitted to the known ASM or, when the category is unknown, to a finite library of candidate priors. A visibility-weighted fitting residual selects the final pose, metric scale, shape, and, when applicable, category (Sec~\ref{sec:admm}). Online computation time of 33--147~ms from input RGB-D to output shape, pose, and class prediction. 
    % \nfcomment{Relate each block or main component in the pipeline figure to their corresponding section in the paper. Do this in the caption and also in the figure.}
    }
    \label{fig:pipeline}
\end{figure*}

\subsection{Problem Formulation}
Given an object region of interest (ROI)\footnote{Note the prior ASM-based pose and shape estimation works~\cite{shi2023optimal,zhang2025shapeicp,shaikewitz2025category} assume known object ROI and correspondences.} in a synchronized RGB-D frame and known camera intrinsics, we seek to estimate the object's rotation $\mathbf R^o\in\mathrm{SO}(3)$, translation $\mathbf t^o\in\mathbb R^3$, positive isotropic scale $s^o$, and ASM shape coefficients $\mathbf c^o\in\mathbb R^K$. The pose and scale define the similarity transformation $\mathbf T^o=(s^o,\mathbf R^o,\mathbf t^o)\in\mathrm{Sim}(3)$. For known-category inference, the corresponding ASM is supplied. Otherwise, a prior is selected from a finite ASM library (constructed in Section~\ref{sec:asm_construction}). Hereafter, we omit the object index $o$ for readability.

A category-level ASM \cite{zhang2025shapeicp} represents the canonical position of template point $i$ of an object as
\begin{equation}
\label{eq:asm-eq}
    \mathbf{p}_i
    =
    \mathbf{b}_i^0
    +
    \sum_{k=1}^K c_k \mathbf{b}_i^k
    =
    \mathbf{b}_i^0 + \mathbf{B}_i\mathbf{c},
\end{equation}
where $\mathbf b_i^0\in\mathbb R^3$ is the mean point position and $\mathbf B_i\in\mathbb R^{3\times K}$ the retained PCA basis vectors representing the primitive shapes of an object dataset. Training shapes are aligned and normalized before PCA so that $s$ controls global metric scale, while $\mathbf c$ parameterizes variation in the normalized shape space. 
% Each PCA mode is scaled by its training-set standard deviation, such that the resulting coefficients have unit variance and admit the isotropic quadratic penalty $\|\mathbf c\|_2^2$.

% A correspondence hypothesis contains a camera-frame point $\hat{\mathbf y}_i\in\mathbb R^3$ and predicted visibility weight $w_i\in[0,1]$ for every template index $i$.
A \textbf{correspondence hypothesis} is one complete ordered set
$\mathcal H=\{(\hat{\mathbf y}_i,w_i)\}_{i=1}^{N}$, where $\hat{\mathbf y}_i\in\mathbb R^3$ is the predicted camera-frame location and $w_i\geq 0$ is the weight of template point $i$. Thus, one hypothesis represents an entire candidate object geometry, including occluded points.
The points are reconstructed from predicted image coordinates and metric depths using the crop transform and camera intrinsics. The shared index is inherited from the common template used to construct the ASM. The same template points are projected into training images to form the correspondence-generator targets described in Section~\ref{sec:complete_correspondence}: image coordinates, relative depth, and visibility.
% \nfcomment{Reader has no idea what the generator targets are, explain how/where they are used in the pipeline and refer to Section \ref{sec:complete_correspondence}.}
Thus, predicted point \(i\) corresponds directly to ASM point \(i\), and no additional correspondence search is required during fitting.
% Because index $i$ is shared with ASM point $i$, no additional correspondence search is required during fitting.

Given $\mathcal H$ the pose, scale, and shape of the object are estimated by solving the following optimization problem
\begin{equation}
    \resizebox{0.98\linewidth}{!}{$
    \begin{aligned}
        \min_{\substack{
        \mathbf{R} \in \mathrm{SO}(3),\,
        \mathbf{t} \in \mathbb{R}^3\\
        s \in \mathbb{R}_{++},\,
        \mathbf{c} \in \mathbb{R}^K}}
        \quad
        & \frac{1}{N}\sum_{i=1}^N
        w_i
        \left\|
        \frac{1}{s}\mathbf{R}^{\top}
        (\hat{\mathbf y}_i-\mathbf t)
        -
        (\mathbf b_i^0+\mathbf B_i\mathbf c)
        \right\|_2^2
        +
        \lambda_{\mathbf c}\|\mathbf c\|_2^2 .
    \end{aligned}$}
\label{eq:prob1}
\end{equation}
where the coefficient $\lambda_{\mathbf c}>0$ controls regularization of the shape coefficients, while predicted visibility weights reduce the contribution of poorly observed points. This optimization problem can be analytically reduced to a nine-dimensional parameter vector and solved in real-time as shown in Section~\ref{sec:admm}.
%We assume $w_i\geq0$ and $\sum_i w_i>0$; together with $\lambda_{\mathbf c}>0$, these conditions ensure that the subsequent translation and coefficient elimination is well defined. Because the residual is evaluated in the canonical frame, Eq.~\eqref{eq:prob1} is not equivalent to a regularized camera-frame least-squares objective when $s$ is optimized.

% \noindent\textbf{Assumptions.}
% We assume (A1) a supplied object ROI, (A2) known camera intrinsics, and (A3) a common template index between model outputs and ASM points. The correspondence predictor does not require a pixel-wise instance mask as input. Dataset-specific ROI and depth preprocessing are described with the corresponding evaluation protocols.

% \noindent\textbf{Inputs and assumptions.}
% We assume a supplied object ROI and known camera intrinsics. The correspondence predictor does not require a pixel-wise instance mask. The common indexing between model outputs and ASM points is constructed offline from the shared template and is therefore part of the representation rather than a test-time input.

\noindent\textbf{Pipeline.}
Fig.~\ref{fig:pipeline} summarizes how the variables above are constructed and used. Section~\ref{sec:asm_construction} first builds the indexed ASM library, providing the mean and basis terms $(\mathbf b_i^0,\mathbf B_i)$ in Eq.~\eqref{eq:asm-eq}. Given an RGB-D ROI, Section~\ref{sec:complete_correspondence} then generates multiple correspondence hypotheses $\mathcal H_h=\{(\hat{\mathbf y}^h_{i},\hat w^h_{i})\}_{i=1}^{N}$ in the same indexed space. Section~\ref{sec:admm} fits each hypothesis to the applicable ASM category prior by solving Eq.~\eqref{eq:prob1}, recovering pose, metric scale, and shape coefficients. Finally, Section~\ref{sec:hypothesis_selection} ranks the valid hypothesis--prior pairs and returns the final estimate.

% \nfcomment{This paragraph should be improved. Explain to the reader what each component does and relate it to the variables you introduce in the problem formulation about. For example, .. ``to obtain $\mathcal H$ we first introduce a fixed point-set indexed representation that is used to construct the ASM library parameterized by $b_i^0,B_i$ from Eq.1, as described in Section III.B. Then, given an unlabeled point set of a query object in a scene we estimate $\mathcal H$ with a novel flow-based generative model as described in Section \ref{sec:complete_correspondence}, etc." This is important since there are so many components in the pipeline, this will help the reader follow it, even if it seems repetitive.}

% \noindent\textbf{Pipeline.}
% As summarized in Fig.~\ref{fig:pipeline}, standardized training shapes define the ASM representation (Section~\ref{sec:asm_construction}), the learned model generates complete indexed correspondence hypotheses (Section~\ref{sec:complete_correspondence}), and each hypothesis is fitted to one or more candidate ASMs (Section~\ref{sec:admm}). The final hypothesis--prior pair is selected using the visibility-weighted fitting residual described in Section~\ref{sec:hypothesis_selection}.
% \input{chapters/method}
\subsection{Active Shape Model Construction}
\label{sec:asm_construction}

\begin{figure}[!t]
    \centering
    \includegraphics[width=1.0\linewidth]
    {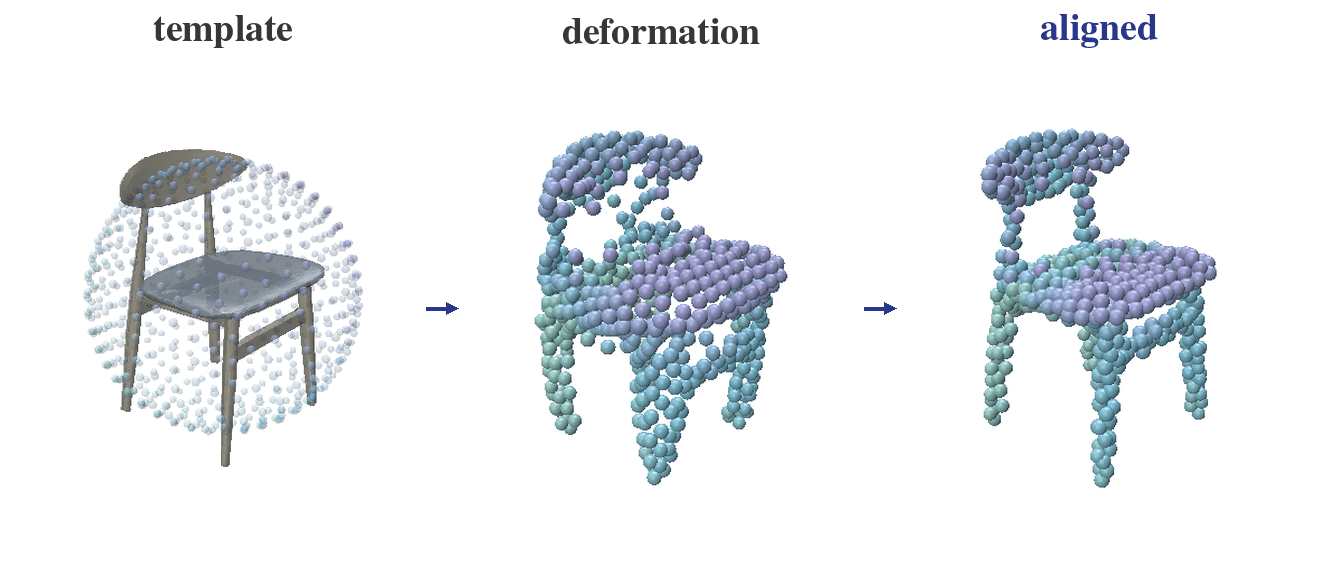}
    \caption{\textbf{Shape standardization.} A common 642-vertex icosphere template is deformed onto each normalized object surface while preserving vertex indices.}
    \label{fig:point_cloud_standardization}
\end{figure}

\subsubsection{Indexed Shape Standardization}
An ASM requires a fixed set of consistently indexed points $N$ across training instances. We construct the library from all 4,385 Omni6DPose CAD models~\cite{Zhang2024Omni6DPoseAB} spanning 156 categories. For each CAD model,
% \nfcomment{How big is this dataset? How many object categories?}
we sample $2^{16}$ surface points only as the dense target surface for deformation. Then, we normalize the shape to a centered unit cube, and deform a common $N=642$ indexed icosphere vertices onto the surface by minimizing the Sinkhorn divergence \cite{feydy2019interpolating} (Fig.~\ref{fig:point_cloud_standardization}). 
% \nfcomment{So N is 642 or $2^{16}$? Define here $N=?$ and explain why you choose that number}.

Vertex indices are preserved during deformation, yielding a consistent template indexing across standardized instances. The resulting points define both the prediction targets and the ASM coordinates, providing approximate cross-instance correspondences for PCA modeling and point-to-point fitting.

\subsubsection{PCA-Based Active Shape Models}
For each category, PCA is applied to the vectorized standardized point sets to obtain a mean shape $\mathbf b^0\in\mathbb R^{N\times3}$ and $K$ retained deformation modes or PCA rank $\mathbf B\in\mathbb R^{N\times3\times K}$. We retain at most 48 modes, which capture $99.9\%$ of the variance or more across all classes; the rank ablation in Appendix~\ref{app:additional_ablations} shows that most task-level gains are already captured by the first few modes.
% \nfcomment{How do you define K? Is it related to the number of categories? Can it be smaller than that number?}
Each mode is scaled by its training-set standard deviation, yielding unit-variance coefficients. The regularization term $\lambda_{\mathbf c}\|\mathbf c\|_2^2$ in Eq.~\eqref{eq:prob1} therefore applies uniform penalization in the normalized PCA coordinate system, corresponding to a scaled Mahalanobis penalty in the original PCA coordinates.
% \nfcomment{How many PCA coordinates do you retain in total? What is the \% of variance they represent (distribution of eigenvalues from PCA)? Be more concreted about these details.}

\subsubsection{Template-Point Projection and Visibility Labeling}
Training targets
% \nfcomment{training targets for what? remind the reader what these training targets will be used for}
for the correspondence generator are obtained by transforming and projecting the standardized template points using the annotated object pose. Each of the $N=642$ template indices is supervised with crop-normalized image coordinates, camera-frame metric depth, and a visibility label.

Our Omni6DPose~\cite{Zhang2024Omni6DPoseAB} model is trained on SOPE--a synthetic dataset--and evaluated on the separately collected ROPE real-world benchmark. No ROPE images are used during training, following Omni6DPose evaluation protocol. For SOPE, the object crop is defined by the projected 3D bounding-box corners and resized to $224\times224$. A template point is labeled as occluded when its projected depth lies more than 1cm behind the ground-truth surface depth. Points that lie outside the image frame, behind the camera, or on invalid depth pixels are marked as unobserved. Coordinate and depth supervision are retained for hidden points, while visibility is predicted separately. Our HouseCat6D\cite{jung2024housecat6d} model is trained independently using its official split, following the mesh- and mask-based visibility procedure defined in its evaluation protocol.

% \nfcomment{For a reader who is not familiar with the Omni6DPose dataset and the HouseCAT6D dataset this last paragraph makes no sense, what is SOPE? do you not train on the ROPE subdataset? I suggest that you include a subsubsection before ``Indexed Shape Standardization" describing the datasets you use for ASM model construction, including \# of categories and number of instances and what all the acronyms mean.}
\section{Complete Correspondence Flow Maps}
\label{sec:complete_correspondence}
The previous section defines the ASM parameters $(\mathbf b_i^0, \mathbf B_i)$ and an indexed template shared across instances. We now introduce a generative flow-based process to predict a distribution over correspondence hypotheses $\mathcal H$ in this same indexed space from an object RGB-D ROI.
% \nfcomment{Add a couple sentences that smoothly segway the reader into this section from the previous. For example, ``Given the learned ASM library parameters $(b_i^0, B_i)$, we now introduce a generative flow-based process to predict correspondence Hypothesis $\mathcal H$ from a query object point set XX.... with $N=624$." Like you have in Section V.}

\subsection{RGB-D Feature Encoding}
\label{sec:encoder}

The correspondence generator conditions on features from the pretrained LingBot-Depth RGB-D encoder \cite{tan2026masked}. The backbone weights remain fixed, while LoRA or DoRA adapters \cite{hu2022lora,liu2024dora} adapt selected projection layers. We evaluate both LoRA and DoRA adaptation; the validation-selected LoRA model is used for the distilled model and subsequent experiments. 
% \nfcomment{Which one do you use LoRA or DoRA or both?}
The predictor receives an RGB-D object crop without a pixel-wise instance mask. Features from layers 4, 8, 12, 16, 20, and 24 are projected to a common token dimension ($D=384$)
% \nfcomment{What is the dimension?}
and combined at each decoder layer using learned softmax-normalized weights.

\subsection{Flow Representation and Decoder}
\label{sec:decoder}

For each template index point $i$, the generated state contains crop-normalized image coordinates and centered relative log-depth
\begin{equation}
\label{eq:revision-3}
\begin{aligned}
    \mathbf{x}_{0,i}
    &=
    \begin{bmatrix}
        u_i & v_i & \delta_i
    \end{bmatrix}^{\top},
    \qquad
    \mathbf{X}_0
    =
    \left\{
        \mathbf{x}_{0,i}
    \right\}_{i=1}^{N},
    \\
    \delta_i
    &=
    \log d_i-\log d_{\mathrm{gm}},
    \qquad
    \log d_{\mathrm{gm}}
    =
    \frac{1}{N}
    \sum_{j=1}^{N}\log d_j
\end{aligned}
\end{equation}
Here, $(u_i,v_i)$ denote crop-normalized image coordinates, $d_i$ is camera-frame depth, and $d_{\mathrm{gm}}$ is the geometric mean depth of the template points. The decomposition $d_i=d_{\mathrm{gm}}\exp(\delta_i)$ separates global camera depth from relative depth structure.
% \nfcomment{How is x related to y? In the next section you start with, given the correspondence set y.. this is confusing. What is the relationship between these variables, or how do you turn $X_t$ into $\hat{y}_i$?}

During training, states follow the linear data-to-noise path
\begin{equation}
\label{eq:revision-4}
    \mathbf{X}_t
    =
    (1-t)\mathbf{X}_0
    +
    t\boldsymbol{\epsilon},
    \qquad
    \boldsymbol{\epsilon}
    \sim
    \mathcal{N}
    \left(
        \mathbf{0},
        \mathbf{I}
    \right)
\end{equation}
Conditioned on $\mathbf X_t$, time $t$, and RGB-D features, the decoder predicts the clean endpoint $\hat{\mathbf X}_0$. A learned query and 2D anchor are assigned to each of the $N=642$ template indices. Six decoder layers alternate point self-attention, RGB-D cross-attention, and anchor refinement. Separate prediction heads output the correspondence state, visibility probability $\hat{\nu}$, and global log-depth $\log\hat d_{\mathrm{gm}}$. After denoising, undoing crop normalization and back-projecting $\mathbf x_{0,i}$ with the camera intrinsics gives $\hat{\mathbf y}_i\in\mathbb R^3$.

% \nfcomment{Is this decoder architecture common or you are proposing it? If the former cite relevant work, if the latter I think you should provide a figure of the decoder. Also where is the decoder represented in Fig. 2? I only see encoder. Is it the flow-matching denoising block?}

\subsection{Training Objective}
\label{sec:training_objective}

We supervise endpoint reconstruction, visibility, and global depth
\begin{equation}
\label{eq:keypoint_detection_loss}
\begin{aligned}
    \mathcal{L}_{\mathrm{endpoint}}
    &=
    \frac{1}{3N}
    \sum_{i=1}^{N}
    \left\lVert
        \widehat{\mathbf{x}}_{0,i}
        -
        \mathbf{x}_{0,i}
    \right\rVert_1,
    \\
    \mathcal{L}_{\mathrm{vis}}
    &=
    \frac{1}{N}
    \sum_{i=1}^{N}
    \operatorname{BCE}
    \left(
        \hat{\nu}_i,
        \nu_i
    \right),
    \\
    \mathcal{L}_{\mathrm{depth}}
    &=
    \left|
        \log\widehat{d}_{\mathrm{gm}}
        -
        \log d_{\mathrm{gm}}
    \right|,
    \\
    \mathcal{L}_{\mathrm{det}}
    &=
    \mathcal{L}_{\mathrm{endpoint}}
    +
    \lambda_{\mathrm{vis}}\mathcal{L}_{\mathrm{vis}}
    +
    \lambda_{\mathrm{depth}}\mathcal{L}_{\mathrm{depth}}
\end{aligned}
\end{equation}
% \nfcomment{Are the v's in $\mathcal{L}_{\mathrm{vis}}$ same as the v in Eq. 3?}

The endpoint loss supervises all template points, including occluded and unobserved targets. Predicted visibility is subsequently used as a soft weight during geometric fitting. 
% We additionally use the auxiliary pose-consistency loss in Appendix~\ref{appx:pose_consistency} to encourage geometric coherence.
% \nfcomment{You should bring the pose-consistent loss text from the Appendix here. This seems like a crucial part in order to make the training converge.}
We additionally use the auxiliary pose-consistency loss. We convert the predicted visibility probabilities into normalized soft weights:
\begin{equation}
    \hat{w}_i
    =
    \frac{
        \hat{\nu}_{i}
    }{
        \sum_{j=1}^{N}
        \hat{\nu}_{j}
    }.
\end{equation}
Let $\hat{\mathbf{y}}_i$ denote the predicted camera-frame keypoint and
$\mathbf{p}_i$ its corresponding canonical keypoint. The similarity pose
induced by the predicted correspondence set is obtained from
\begin{equation}
\begin{aligned}
    \left(
        \hat{s},
        \hat{\mathbf{R}},
        \hat{\mathbf{t}}
    \right)
    =
    \underset{\substack{
        s\in\mathbb{R}_{++},\,
        \mathbf{R}\in\mathrm{SO}(3)\\
        \mathbf{t}\in\mathbb{R}^{3}
    }}{\operatorname{argmin}}
    \sum_{i=1}^{N}
    \hat{w}_i
    \left\lVert
        \hat{\mathbf{y}}_i
        -
        s\mathbf{R}\mathbf{p}_i
        -
        \mathbf{t}
    \right\rVert_2^2.
\end{aligned}
\end{equation}
We supervise the resulting rotation and translation using
\begin{equation}
\begin{aligned}
    \mathcal{L}_{\mathrm{pose}}
    ={}&
    (1-t)^2
    \Bigg[
        1
        -
        \frac{
            \operatorname{tr}
            \left(
                \hat{\mathbf{R}}^{\top}
                \mathbf{R}_{\mathrm{gt}}
            \right)
            -1
        }{2}
        \\
        &\quad+
        \frac{\lambda_{\mathrm{trans}}}{3}
        \left\lVert
            \boldsymbol{\phi}
            \left(
                \hat{\mathbf{t}},
                \hat{d}_{\mathrm{gm}}
            \right)
            -
            \boldsymbol{\phi}
            \left(
                \mathbf{t}_{\mathrm{gt}},
                d_{\mathrm{gm}}
            \right)
        \right\rVert_1
    \Bigg],
\end{aligned}
\label{eq:keypoint_pose_loss}
\end{equation}
where $\boldsymbol{\phi}(\mathbf{t},d)$ contains the crop-normalized
projection of $\mathbf{t}$ and the relative log-depth $\log(t_z/d)$. The
factor $(1-t)^2$ reduces the influence of pose estimates obtained from highly
corrupted flow states. Metric scale is supervised separately through the
global-depth loss.
The complete training objective is
\[
\mathcal L_{\mathrm{det}}
+
\lambda_{\mathrm{pose}}\mathcal L_{\mathrm{pose}}
\]
For $t>0$, the endpoint predictor defines the velocity field
\begin{equation}
\mathbf v_\theta(\mathbf X_t,t;\mathcal C)
=
\frac{
\mathbf X_t-
\widehat{\mathbf X}_{0,\theta}(\mathbf X_t,t;\mathcal C)
}{t},
\qquad t>0,
\label{eq:endpoint-velocity}
\end{equation}
where $\mathcal C$ denotes the RGB-D condition. Integrating this field from the noise endpoint toward $t=0$ produces a correspondence hypothesis. At $t=0$, each endpoint $\hat{\mathbf x}_{0,i}$ is converted to $\hat{\mathbf y}_i$ by metric-depth reconstruction and back-projection, while the visibility head provides $\hat{w}_i$. Together these form $\mathcal H=\{(\hat{\mathbf y}_i,\hat{w}_i)\}_{i=1}^N$.

% \nfcomment{how? show how you go from the final $x_t$ after velocity field integration and the correspendce set y. Also what happened to the visibility weights w? Aren't these part of the hypothesis too? Too many details are glossed over.}

\subsection{Inference and Distillation}
\label{sec:inference}
At inference, multiple correspondence hypotheses are obtained from independent Gaussian initializations by integrating the learned velocity field $\mathbf v_\theta$ with Euler steps. To reduce the cost of this iterative denoising, we distill the 10-step teacher into a target-time-conditioned flow map that generates hypotheses in only two or four steps. The four-step model is used by default since it preserves pose accuracy while reducing decoder evaluations (Table~\ref{tab:comparison_and_adaptation}).
% \nfcomment{Bring the text in Appendix C to this section. This seems like a critical component that allows for the real-time inference. How fast is the distilled map vs. the original flow-map?}
We distill the trained model into a flow map conditioned on the current time $t$ and target time $t'<t$~\cite{sabour2026align}. The RGB-D condition remains fixed and is omitted from the notation. The
student transition is
\begin{equation}
\begin{aligned}
    \tilde{\mathbf{X}}_{t',\psi}
    \left(
        \mathbf{X}_t,t,t';\mathcal C
    \right)
    =
    \mathbf{X}_t
    +
    \frac{t'-t}{t}
    \left[
        \mathbf{X}_t
        -
        \hat{\mathbf{X}}_{0,\psi}
        \left(
            \mathbf{X}_t,t,t';\mathcal C
        \right)
    \right].
\end{aligned}
\label{eq:student_flow_map}
\end{equation}
The student is initialized from the trained model with a zero-initialized
target-time embedding. The RGB-D encoder remains frozen, and only the decoder
is optimized. For each interval $[t',t]$, the frozen teacher is integrated from
$\mathbf{X}_t$ to $\hat{\mathbf{X}}_{t',\theta}$ using four uniform
substeps. The student matches the resulting average velocity:
\begin{equation}
\begin{aligned}
    \mathcal{L}_{\mathrm{rollout}}
    =
        \mathcal{L}_{\kappa}
        \Bigg(
            \frac{
                \tilde{\mathbf{X}}_{t',\psi}
                -
                \hat{\mathbf{X}}_{t',\theta}
            }{t'-t}
        \Bigg) ,
\end{aligned}
\label{eq:rollout_distill}
\end{equation}
where $\mathcal{L}_{\kappa}$ is the pseudo-Huber penalty~\cite{song2024improved} with
$\kappa=0.05$ averaged over samples,
keypoints, and state coordinates. We also enforce agreement between direct and composed transitions~\cite{frans2025one}. Let
$t''=t'+\eta(t-t')$, where
$\eta\sim\mathcal{U}(0.35,0.65)$, and define
\begin{equation}
\begin{aligned}
    \tilde{\mathbf{X}}_{t',\psi}'
    \left(
        \mathbf{X}_t,t,t',t''; \mathcal{C}
    \right)
    &=
    \tilde{\mathbf{X}}_{t',\psi}
    \left(
        \tilde{\mathbf{X}}_{t'',\psi}
        \left(
            \mathbf{X}_t,t,t''; \mathcal{C}
        \right),
        t'',t';\mathcal{C}
    \right).
\end{aligned}
\end{equation}
The composition loss is
\begin{equation}
    \mathcal{L}_{\mathrm{comp}}
    =
    \mathcal{L}_{\kappa}
    \left(
        \tilde{\mathbf{X}}_{t',\psi}
        -
        \operatorname{sg}
        \left[
        \tilde{\mathbf{X}}_{t',\psi}'
        \right]
    \right),
\label{eq:composition_consistency}
\end{equation}
where $\operatorname{sg}$ denotes stop-gradient. Transitions terminating at $t'=0$ receive additional endpoint supervision:
\begin{equation}
\begin{aligned}
    \mathcal{L}_{\tilde{\mathbf{X}}_{0}}
    &=
    \mathcal{L}_{\kappa}
    \left(
        \tilde{\mathbf{X}}_{0,\psi}
        -
        \mathbf{X}_0
    \right),
    \mathcal{L}_{\hat{\mathbf{X}}_0}
    =
    \mathcal{L}_{\kappa}
    \left(
        \hat{\mathbf{X}}_{0,\psi}
        -
        \mathbf{X}_0
    \right).
\end{aligned}
\end{equation}
The complete distillation objective is
\begin{equation}
\begin{aligned}
    \mathcal{L}_{\mathrm{distill}}
    ={}&
    \mathcal{L}_{\mathrm{rollout}}
    +
    0.25\mathcal{L}_{\mathrm{comp}}
    +
    0.25\mathcal{L}_{\tilde{\mathbf{X}}_{0}}
    +
    0.05\mathcal{L}_{\hat{\mathbf{X}}_0}
    \\& +
    10^{-3}\mathcal{L}_{\mathrm{vis}}
    +
    10^{-2}\mathcal{L}_{\mathrm{depth}}.
\end{aligned}
\label{eq:distill_total}
\end{equation}
The final two terms preserve visibility and global-depth prediction during
distillation. Training intervals include continuous transitions and those
used by the one-, two-, and four-step schedules, with additional samples
ending at $t'=0$. At inference, the distilled flow map is composed over two
or four steps from $t=1$ to $t=0$.

The distilled model is initialized from the selected LoRA model and trained for 10 epochs using AdamW with batch size 64, learning rate $3\times10^{-4}$, and weight decay $10^{-4}$; only the student flow decoder is updated. We reconstruct $\hat{\mathbf y}_i$ from $(\hat u_i,\hat v_i,\hat\delta_i,\hat d_{\rm gm})$ and pass $\{(\hat{\mathbf y}_i,\hat{w}_i)\}$ directly to Eq.~\ref{eq:prob1}.
% \nfcomment{How? Be concrete.}

% \nfcomment{Overall, this section is too thin. You gloss over critical design choices or push them to the appendix. Please re-write this section following my suggestions.}
\section{Scale-Aware ASM Fitting and Selection}
\label{sec:admm}

Given the 3D points $\{\hat{\mathbf{y}}_i\}_{i=1}^{642}$ representing the complete correspondence set for hypothesis $\mathcal H$, we solve Eq.~\eqref{eq:prob1} to obtain pose, scale, and shape of the object of interest. The coupling between rotation and metric scale makes Eq.~\eqref{eq:prob1} nonconvex. Semidefinite relaxations~\cite{Holmes2024OnSR} can provide global-optimality certificates, but are too costly for the large batches considered in this work.
%The problem is a nonconvex quadratically constrained quadratic program (QCQP) because of the rotation constraint $\mathbf{R} \in \mathrm{SO}(3)$. Semidefinite relaxations~\cite{Holmes2024OnSR} can provide global-optimality certificates, but are too costly for the large batches considered here.
%We therefore reparameterize Eq.~\eqref{eq:prob1} using the scaled-rotation manifold $\mathcal{M}=\{s\mathbf{R}\mid s>0,\ \mathbf{R}\in\mathrm{SO}(3)\}$, which reduces the problem to a quadratic program. We then solve the reduced problem with consensus ADMM~\cite{Huang2016ConsensusADMMFG, huang2023convergence} tailored to Sim(3) estimation.
We therefore introduce the inverse metric scale $\alpha=s^{-1}$ and use the scaled-rotation set \begin{equation} \mathcal{M} = \left\{ \alpha\mathbf{Q} \mid \alpha>0,\; \mathbf{Q}\in\mathrm{SO}(3) \right\}, \end{equation} which reduces the optimization problem to a quadratically constrained quadratic program (QCQP), as shown in Section~\ref{sec:reduced}. 

We then solve the reduced problem, Eq.~\eqref{eq:admm_full}, with consensus ADMM~\cite{Huang2016ConsensusADMMFG, huang2023convergence} tailored to Sim(3) estimation, as described in Section~\ref{sec:batch-admm}. Note that these steps are done for a single hypothesis correspondence set $\mathcal H$. Section~\ref{sec:hypothesis_selection} describes how multiple of such hypotheses are ranked with a visibility-weighted score.

\subsection{Reduced Pose–Shape Formulation}
\label{sec:reduced}

We begin by reparameterizing Eq.~\eqref{eq:prob1} to decouple the optimization variables.
%Let $\mathbf{\Omega} = \frac{1}{s}\mathbf{R}^\top$ and $\boldsymbol{\tau} = \frac{1}{s}\mathbf{R}^\top \mathbf{t}$, yielding the following objective:
% \begin{equation}
%     \begin{aligned}
%         \min_{\substack{ \mathbf{\Omega} \in \mathcal{M}, \boldsymbol{\tau} \in \mathbb{R}^3, \\ \mathbf{c} \in \mathbb{R}^K }} 
%         & \frac{1}{N} \sum_{i=1}^N w_i \left\| \mathbf{\Omega}\hat{\mathbf{y}}_i - (\mathbf{b}_i^0 + \mathbf{B}_i \mathbf{c}) - \boldsymbol{\tau} \right\|_2^2 + \lambda_\mathbf{c} \|\mathbf{c}\|_2^2
%     \end{aligned}
% \label{eq:admm_full}
% \end{equation}
Let
\begin{equation}
    \alpha=s^{-1},
    \qquad
    \mathbf{\Omega}=\alpha\mathbf{R}^{\top},
    \qquad
    \boldsymbol{\tau}=\alpha\mathbf{R}^{\top}\mathbf{t},
\end{equation}
where $\alpha>0$ denotes inverse metric scale. Since
$\mathbf{R}^{\top}\in\mathrm{SO}(3)$, we have
$\mathbf{\Omega}\in\mathcal{M}$. The objective becomes
\begin{equation}
    \begin{aligned}
        \min_{\substack{
        \mathbf{\Omega} \in \mathcal{M},
        \boldsymbol{\tau} \in \mathbb{R}^3, \\
        \mathbf{c} \in \mathbb{R}^K}}
        \quad&
        \frac{1}{N}
        \sum_{i=1}^N
        w_i
        \left\|
        \mathbf{\Omega}\hat{\mathbf{y}}_i
        -
        (\mathbf{b}_i^0+\mathbf{B}_i\mathbf{c})
        -
        \boldsymbol{\tau}
        \right\|_2^2
        +
        \lambda_{\mathbf c}\|\mathbf{c}\|_2^2 .
    \end{aligned}
\label{eq:admm_full}
\end{equation}
The formulation remains in the canonical frame, since optimizing metric scale makes the corresponding regularized camera-frame objective different. For fixed $\mathbf{\Omega}$, Eq.~\eqref{eq:admm_full} is jointly quadratic in $\mathbf c$ and $\boldsymbol{\tau}$. Assuming $\sum_i w_i>0$ and $\lambda_{\mathbf c}>0$, both variables can be eliminated analytically as functions of $\mathbf{\Omega}$. This yields a compact nonconvex quadratic program in the vectorized scaled rotation $\boldsymbol{\omega}=\operatorname{vec}(\mathbf{\Omega})$.

\begin{proposition}[QP Formulation] 
\label{prop:qp}
The pose--shape problem in Eq.~\eqref{eq:admm_full} is equivalent to the following nonconvex quadratic program on $\mathcal{M}$:
\begin{equation}
    \begin{aligned}
        \min_{\boldsymbol{\omega} \in \mathbb{R}^9} \quad & \frac{1}{2}\boldsymbol{\omega}^\top \mathbf{H} \boldsymbol{\omega} + \mathbf{g}^\top \boldsymbol{\omega} \\
        \text{s.t.} \quad & \mathrm{mat}(\boldsymbol{\omega}) \in \mathcal{M}
    \end{aligned}
\label{eq:qcqp}
\end{equation}
where the Hessian matrix $\mathbf{H}(\theta)\in\mathbb{R}^{9\times 9}$ and linear gradient vector $\mathbf{g}(\theta)\in\mathbb{R}^{9}$ are the quadratic and linear terms defined in Eq.~\eqref{eq:qp-terms} from Appendix~\ref{appx:closed_form}. These are computed as linear weighted functions of $\theta = \{\lambda_{\mathbf{c}}, (w_i,\hat{\mathbf{y}}_i,\mathbf{b}_i^0,\mathbf{B}_i)_{i=1}^{N}\}$.\\ \textit{Proof:} See full derivation in Appendix~\ref{appx:closed_form}. \hfill$\blacksquare$
\end{proposition}

\subsection{Batched ADMM Solver}
\label{sec:batch-admm}
% We solve Eq.~\eqref{eq:qcqp}with an ADMM scheme inspired by ReLU-QP~\cite{Bishop2023ReLUQPAG}. Introducing splitting variables $\bar{\boldsymbol{\omega}}$ and $\mathbf{z}$ decouples the quadratic objective from the scaled-rotation constraint $\mathcal{M}=\mathbb{R}_{++}\times\mathrm{SO}(3)$, giving
To solve Eq.~\eqref{eq:qcqp} in a computationally efficient manner, we introduce an ADMM scheme inspired by
ReLU-QP~\cite{Bishop2023ReLUQPAG}. For numerical projection, we use
the closure of the positive scaled-rotation set,
\begin{equation}
    \overline{\mathcal{M}}
    =
    \left\{
    \alpha\mathbf{Q}
    \mid
    \alpha\geq0,\;
    \mathbf{Q}\in\mathrm{SO}(3)
    \right\}.
\end{equation}
To separate the quadratic objective from the nonconvex scaled-rotation constraint, we introduce an unconstrained consensus variable $\bar{\boldsymbol{\omega}}$ and a constrained copy $\mathbf z$. The equality constraints force these variables to agree with the scaled-rotation estimate at convergence.
% Introducing splitting variables $\bar{\boldsymbol{\omega}}$ and
% $\mathbf{z}$ decouples the quadratic objective from the
% scaled-rotation constraint, giving
\begin{equation}
\label{eq:admm-qp}
    \begin{aligned}
        \min_{\boldsymbol{\omega},\bar{\boldsymbol{\omega}},\mathbf{z}}
        \quad &
        \frac{1}{2}\bar{\boldsymbol{\omega}}^\top
        \mathbf{H}\bar{\boldsymbol{\omega}}
        +\mathbf{g}^\top\bar{\boldsymbol{\omega}}
        +\mathbb{I}_{\overline{\mathcal{M}}}(\mathbf{z}) \\
        \text{s.t.}\quad &
        \bar{\boldsymbol{\omega}}=\mathbf{z},
        \qquad
        \boldsymbol{\omega}=\bar{\boldsymbol{\omega}} .
    \end{aligned}
\end{equation}
The corresponding augmented Lagrangian is
\begin{equation}
    \begin{aligned}
        L_{\rho,\sigma}
        &=
        \frac{1}{2}\bar{\boldsymbol{\omega}}^\top
        \mathbf{H}\bar{\boldsymbol{\omega}}
        +\mathbf{g}^\top\bar{\boldsymbol{\omega}}
        +\mathbb{I}_{\mathcal{M}}(\mathbf{z})
        +\frac{\sigma}{2}
        \|\bar{\boldsymbol{\omega}}-\boldsymbol{\omega}\|_2^2 \\
        &\quad
        +\boldsymbol{\mu}^\top
        (\bar{\boldsymbol{\omega}}-\boldsymbol{\omega})
        +\frac{\rho}{2}
        \|\bar{\boldsymbol{\omega}}-\mathbf{z}\|_2^2
        +\boldsymbol{\lambda}^\top
        (\bar{\boldsymbol{\omega}}-\mathbf{z}),
    \end{aligned}
\end{equation}
where $\boldsymbol{\mu}$ and $\boldsymbol{\lambda}$ are the dual variables associated with $\bar{\boldsymbol{\omega}}=\boldsymbol{\omega}$ and $\bar{\boldsymbol{\omega}}=\mathbf z$, respectively, and $\sigma,\rho\in\mathbb R_{++}$ are the corresponding penalty parameters.
%where $\sigma,\rho\in\mathbb{R}_{++}$ are the consensus and scaled-rotation penalty parameters, respectively.
As in ReLU-QP, the consensus updates yield $\boldsymbol{\mu}=\mathbf{0}$ and $\bar{\boldsymbol{\omega}}=\boldsymbol{\omega}$ after the first iteration, allowing these variables to be eliminated from the iterative solver~\cite{Bishop2023ReLUQPAG}. The resulting updates are
\begin{equation}
\begin{aligned}
\boldsymbol{\omega}^{t+1}
&=
\widetilde{\mathbf{H}}_t^{-1}
\left(
-\mathbf{g}
+\sigma\boldsymbol{\omega}^{t}
+\rho_t\mathbf{z}^{t}
-\boldsymbol{\lambda}^{t}
\right), \\
\mathbf{z}^{t+1}
&=
\mathcal{P}_{\overline{\mathcal{M}}}
\left(
\boldsymbol{\omega}^{t+1}
+\rho_t^{-1}\boldsymbol{\lambda}^{t}
\right), \\
\boldsymbol{\lambda}^{t+1}
&=
\boldsymbol{\lambda}^{t}
+
\rho_t
\left(
\boldsymbol{\omega}^{t+1}-\mathbf{z}^{t+1}
\right),
\label{eq:admm-updates}
\end{aligned}
\end{equation}
where
\begin{equation}
\widetilde{\mathbf{H}}_t^{-1}
=
\left(
\mathbf{H}+(\sigma+\rho_t)\mathbf{I}
\right)^{-1}.
\end{equation}
We do not form the inverse explicitly; instead, we compute each update using batched Cholesky factorization and triangular solves. We compute $\boldsymbol{\omega}^{t+1}$ using a batched Cholesky factorization $\widetilde{\mathbf{H}}_t = \mathbf{L}_t \mathbf{L}_t^\top$. We first solve for the intermediate vector $\mathbf{q}_t$ via forward-substitution:
\begin{equation}
\mathbf{L}_t \mathbf{q}_t = -\mathbf{g}
+\sigma\boldsymbol{\omega}^{t}
+\rho_t\mathbf{z}^{t}
-\boldsymbol{\lambda}^{t},
\end{equation}
and then obtain $\boldsymbol{\omega}^{t+1}$ via back-substitution:
\begin{equation}
\mathbf{L}_t^\top \boldsymbol{\omega}^{t+1} = \mathbf{q}_t.
\end{equation} 
% \nfcomment{Can you write explicitly this batched Cholesky solve? If short place here, if long place in appendix.}

The operator
$\mathcal{P}_{\overline{\mathcal{M}}}(\mathbf{u})$ in Eq.~\eqref{eq:admm-updates} projects onto
$\overline{\mathcal{M}}$. For
$\mathbf{Z}=\operatorname{mat}(\mathbf{u})$, we recover the closest
proper rotation $\widehat{\mathbf{R}}$ using Horn's quaternion
eigensystem~\cite{horn1987closed} and apply the exact Euclidean projection
\begin{equation}
\alpha^\star
=
\max\left(
0,\,
\frac{
\langle
\widehat{\mathbf{R}},
\mathbf{Z}
\rangle_F
}{3}
\right),
\qquad
\Pi_{\overline{\mathcal{M}}}(\mathbf{Z})
=
\alpha^\star\widehat{\mathbf{R}}.
\label{eq:scaled_rotation_projection}
\end{equation}
Horn's formulation is equivalent to the proper-rotation Procrustes
solution while avoiding a batched $3\times3$ SVD.

Further, we adapt $\rho_t$ from Eq.~\eqref{eq:admm-updates} to balance feasibility and stationarity. The initial penalty is scaled by the quadratic curvature,
\begin{equation}
\rho_0
=
\max\left(
\rho_{\mathrm{base}}
\operatorname{median}_{i}(H_{ii}),
\;\rho_{\min}
\right),
\end{equation}
where $\rho_{\mathrm{base}}=3\times10^{-2}$ and $\rho_{\min}=10^{-5}$. Using the normalized residuals, with $\epsilon=10^{-12}$ for numerical stabilization,
\begin{equation}
\widehat r_{\mathrm p}^{\,t}
=
\frac{
\|\boldsymbol{\omega}^{t}-\mathbf{z}^{t}\|_2
}{
\max(
\|\boldsymbol{\omega}^{t}\|_\infty,
\|\mathbf{z}^{t}\|_\infty,
\epsilon)
},
\end{equation}
\begin{equation}
\widehat r_{\mathrm d}^{\,t}
=
\frac{
\|\mathbf{H}\boldsymbol{\omega}^{t}
+\mathbf{g}
+\boldsymbol{\lambda}^{t}\|_\infty
}{
\max(
\|\mathbf{H}\boldsymbol{\omega}^{t}\|_\infty,
\|\mathbf{g}\|_\infty,
\|\boldsymbol{\lambda}^{t}\|_\infty,
\epsilon)
},
\end{equation}
we update
\begin{equation}
\rho_{t+1}
=
\operatorname{clip}
\left(
\rho_t
\sqrt{
\frac{
\widehat r_{\mathrm p}^{\,t}
}{
\max(\widehat r_{\mathrm d}^{\,t},\epsilon)
}
},
\;10^{-3}\rho_0,
\;10\rho_0
\right).
\label{eq:rho_adaptation}
\end{equation}
The penalty therefore increases when feasibility error dominates and decreases when stationarity error dominates.

We use $\lambda_c=\sigma=10^{-5}$ and 20 ADMM iterations for inference. The penalty is adapted every three iterations starting at $t=3$. Among all iterations, we return the feasible $\mathbf{z}^{t}$ with the lowest original quadratic objective. Appendix~\ref{appx:admm_formulation} provides the full projection derivation.

\subsection{Global Optimality Certification}
To certify that the ADMM solution $\boldsymbol{\omega}^*$ to Eq.~\eqref{eq:admm-qp} is the global optimum, we utilize Semidefinite Programming (SDP) duality~\cite{Holmes2024OnSR}. Specifically, we form the SDP relaxation of Eq.~\eqref{eq:qcqp}; if the corresponding dual slack matrix is positive semidefinite and the stationarity conditions are satisfied, $\boldsymbol{\omega}^*$ is certified as a global optimum of the reduced geometric fitting problem.

\begin{proposition}[SDP Relaxation] 
\label{prop:sdp_relaxation}
The following semidefinite program (SDP) is a convex relaxation of Eq.~\eqref{eq:qcqp}
\begin{equation}
    \begin{aligned}
        \min_{\mathbf{W}\succeq \mathbf{0}} \quad & \left\langle \mathbf{Q}, \mathbf{W} \right\rangle \\
        \text{s.t.} \quad & \langle \mathbf{A}_i, \mathbf{W} \rangle = a_i, \quad i=1,\dots,7
    \end{aligned}
\label{eq:sdp_relaxation_main}
\end{equation}
where $\mathbf{W} = \mathbf{w}\mathbf{w}^\top \in \mathbb{S}_{+}^{10}$ is a lifted state with $\mathbf{w} = [\boldsymbol{\omega}^\top, 1]^\top$; $\mathbf{Q}$ is a homogenized cost matrix and $\mathbf{A}_i$'s are manifold enforcing matrices. $\langle \mathbf{A}, \mathbf{B} \rangle = \text{tr}(\mathbf{A}^\top \mathbf{B})$ denotes the Frobenius inner product. If the rank-1 constraint on $\mathbf{W}$ is relaxed to the convex set $\mathbf{W} \succeq \mathbf{0}$, then the solution $\mathbf{w}^*$ is certified as the global minimizer if there exists a multiplier vector $\boldsymbol{\eta} \in \mathbb{R}^7$ such that the dual slack matrix $\mathbf{S} = \mathbf{Q} - \sum \eta_i \mathbf{A}_i$ is positive semidefinite ($\mathbf{S}\mathbf{w}^* = 0, \mathbf{S} \succeq \mathbf{0}$). By solving $\boldsymbol{\eta}$ using a least-squares fit on the stationarity conditions, we obtain:
\begin{equation}
    \boldsymbol{\eta} = \mathbf{M}^\dagger (\mathbf{Q} \mathbf{w}^*)
\end{equation}
for $\mathbf{M} = [\mathbf{A}_1 \mathbf{w}^*, \dots, \mathbf{A}_7 \mathbf{w}^*] \in \mathbb{R}^{10 \times 7}$. We compute candidate multipliers by least
squares and declare the fit numerically certified only if primal feasibility, stationarity/complementarity, and dual-slack positive semidefiniteness pass the specified tests.\\
\textit{Proof:} See Appendix~\ref{appx:sdp_kkt}. \hfill$\blacksquare$
\end{proposition}

% \nfcomment{What about the global optimality certificate? This should be mentioned HERE not in the evaluation section. You don't need to include Appendix F here but at least highlight it like.. "We certify the global optimality of the solution $\omega^*$ to QCQP by relaxing it into a SDP and showing relaxed SDP solution naturally turns out to be rank-on."}

\subsection{Hypothesis and Prior Selection}
\label{sec:hypothesis_selection}
% \nfcomment{This section is so reduced that reader has no idea what you are talking about, what is the prior? how do you generate the prior? what is h? How is this related to $\mathcal H$? Is it an instance of it? What is the index $j$ how many priors do we have? This is also a critical component that is not clearly defined. Bring back Appendix G here. You could keep the depth-consistency diagnostic in appendix, but mention it here and why it is needed.}

% For each valid hypothesis--prior pair $(h,j)$, we define $\gamma_{\mathrm{ASM}}(h,j)$ as the visibility-weighted mean Euclidean distance between the generated and ASM-fitted camera-frame points (Appendix~\ref{app:selection_details}). This ranking score is distinct from the canonical-frame objective used for geometric fitting. The final hypothesis--prior pair is selected as
% \begin{equation}
%     \begin{aligned}
%     (h^\star,j^\star)
%     =
%     \underset{(h,j)\in\mathcal{C}_{\mathrm{valid}}}
%     {\operatorname{argmin}}
%     \;
%     \gamma_{\mathrm{ASM}}(h,j)
%     \end{aligned}
% \label{eq:hypothesis_prior_selection}
% \end{equation}
% Invalid or nonfinite fits are excluded from $\mathcal{C}_{\mathrm{valid}}$. For known-category inference, the prior index $j$ is fixed.

The generator produces $H$ correspondence hypotheses
$\{\mathcal H_h\}_{h=1}^{H}$, where
$\mathcal H_h=\{(\hat{\mathbf y}_i^h,\hat{w}_i^h)\}_{i=1}^{N}$.
Each hypothesis represents one complete camera-frame interpretation of the
object. The index $j$ denotes an ASM prior from the offline library constructed
in Section~\ref{sec:asm_construction}; in our Omni6DPose setting, the library
contains $C=156$ category priors.

Before fitting, each hypothesis optionally receives one sensor-assisted global-depth refinement using visible points with valid RGB-D measurements.
For hypothesis $h$, let $\hat d_{\mathrm{gm}}^h$ denote its predicted global depth. For visible points with valid sensor depth, collected in $\mathcal I_h$, we estimate
\begin{equation}
\begin{aligned}
    \hat{d}_{\mathrm{gm}}^{\mathrm{obs},h}
    =
    \operatorname{wmed}_{i\in\mathcal I_h}
    \left(
        \frac{
            \mathcal D(\hat u_i^h,\hat v_i^h)
        }{
            \exp(\hat{\delta}_i^h)
        };
        \hat{w}_i^h
    \right),
    \hat{d}_{\mathrm{gm}}^{\mathrm{ref},h}
    =
    \sqrt{
        \hat d_{\mathrm{gm}}^h \hat d_{\mathrm{gm}}^{\mathrm{obs},h}
    }.
\end{aligned}
\label{eq:depth_refinement}
\end{equation}
The first term recovers a per-point estimate of the global depth by removing
the predicted relative-depth factor, then aggregates these estimates robustly
with a visibility-weighted median. The result is finally fused with the
model-predicted global depth through their geometric mean. Here,
$\mathcal D$ is the observed depth image and the weighted median uses
$\hat{w}_i^h$. The refined depth $\hat{d}_{\mathrm{gm}}^{\mathrm{ref},h}$ replaces $\hat d_{\mathrm{gm}}^h$ during
metric reconstruction before ASM fitting. We use one refinement step by
default.

We then fit
$\mathcal H_h$ to prior $j$ using Eq.~\eqref{eq:prob1}, yielding pose, metric
scale, shape coefficients, and fitted camera-frame points
$\{\mathbf y_i^{h,j}\}_{i=1}^{N}$.
We rank each valid hypothesis--prior pair using the visibility-weighted mean
fitting residual
\begin{equation}
    \gamma_{\mathrm{ASM}}(h,j)
    =
    \frac{
        \sum_{i=1}^{N}
        \hat{w}_i^h
        \left\|
            \hat{\mathbf y}_i^h-\mathbf y_i^{h,j}
        \right\|_2
    }{
        \sum_{i=1}^{N} \hat{w}_i^h
    }.
\label{eq:deployed_selection}
\end{equation}
This score is evaluated in the camera frame and is used only for candidate
ranking; it is distinct from the canonical-frame objective optimized during
ASM fitting. Invalid or nonfinite fits are discarded, and the final estimate is
selected by
\begin{equation}
    (h^\star,j^\star)
    =
    \operatorname{argmin}_{(h,j)\in\mathcal C_{\mathrm{valid}}}
    \gamma_{\mathrm{ASM}}(h,j).
\label{eq:hypothesis_prior_selection}
\end{equation}
For known-category inference, $j$ is fixed, and selection is performed only
over the $H$ correspondence hypotheses. When the category is unknown, all
$H\times C$ hypothesis--prior pairs are evaluated jointly.

\begin{table*}[!t]
    \caption{Category-level pose estimation on ROPE and ASM-6D adaptation and distillation variants.}
    \label{tab:comparison_and_adaptation}
    \centering
    \setlength{\tabcolsep}{2.5pt}
    \renewcommand{\arraystretch}{1.00}
    \footnotesize
    
    \begin{tabular*}{\textwidth}{
    @{\extracolsep{\fill}}l|ccc|cccc|cc@{}}
    \toprule
    \toprule
    \multirow{2}{*}{Method}
    & \multicolumn{3}{c|}{AUC $\uparrow$}
    & \multicolumn{4}{c|}{VUS $\uparrow$}
    & \multicolumn{2}{c}{Mean error $\downarrow$} \\
    & IoU$_{25}$ & IoU$_{50}$ & IoU$_{75}$
    & $5^{\circ}$/2\,cm & $5^{\circ}$/5\,cm
    & $10^{\circ}$/2\,cm & $10^{\circ}$/5\,cm
    & Rot. ($^{\circ}$) & Trans. (cm) \\
    \hline
    
    \multicolumn{10}{l}{\textit{Prior work}} \\
    NOCS~\cite{wang2019normalized}
    & 0.0 & 0.0 & 0.0
    & 0.0 & 0.0 & 0.0 & 0.0
    & -- & -- \\
    
    SGPA~\cite{chen2021sgpa}
    & 10.5 & 2.0 & 0.0
    & 4.3 & 6.7 & 9.3 & 15.0
    & 60.2 & 2.81\\
    
    IST-Net~\cite{liu2023net}
    & 28.7 & 10.6 & 0.5
    & 2.0 & 3.4 & 5.3 & 8.8
    & 78.4 & 3.40 \\
    
    HS-Pose~\cite{Zheng_2023_CVPR}
    & 31.6 & 13.6 & 1.1
    & 3.5 & 5.3 & 8.4 & 12.7
    & 63.3 & 3.02 \\
    
    GenPose~\cite{zhang2024generative}
    & -- & -- & --
    & 6.6 & 9.6 & 13.1 & 19.3
    & 42.2 & 2.05 \\
    
    GenPose++~\cite{Zhang2024Omni6DPoseAB}
    & 39.0 & 19.1 & 2.0
    & 10.0 & 15.1 & 19.5 & 29.4
    & 30.7 & 1.28 \\
    
    RFMPose$^\dagger$~\cite{ouyang2026rfmpose}
    & 42.1 & 21.0 & 2.2
    & 10.4 & 15.7 & 21.0 & 30.8
    & -- & -- \\
    \hline
    
    \multicolumn{10}{l}{\textit{ASM-6D adaptation and distillation}} \\
    % Direct Regression
    % & 27.87 & 11.58 & 0.93
    % & 4.38 & 6.81 & 10.88 & 16.98
    % & 39.33 & 1.73 & -- \\
    
    Flow Matching (frozen)
    & 46.67 & 28.09 & 5.82
    & 11.65 & 16.23 & 23.07 & 32.57
    & 30.81 & 1.22 \\
    
    Flow Matching (DoRA)
    & 48.29 & 29.98 & 6.83
    & \cellcolor{blue!5}\underline{15.01} & 20.90 & 27.30 & 38.46
    & 23.54 & 1.21 \\
    
    Flow Matching (LoRA)
    & 48.73 & 30.39 & 6.92
    & 14.65 & 20.36 & 27.27 & 38.31
    & \cellcolor{blue!5}\underline{23.44} & \cellcolor{blue!5}\underline{1.17} \\
    
    \enspace + Distill ($n_{\mathrm{step}}=2$)
    & \cellcolor{blue!5}\underline{48.82} & \cellcolor{blue!5}\underline{30.93} & \cellcolor{green!10}\textbf{7.71}
    & \cellcolor{green!10}\textbf{15.46} & \cellcolor{blue!5}\underline{21.51}
    & \cellcolor{blue!5} \underline{27.70} & \cellcolor{blue!5}\underline{38.90}
    & 23.85 & \cellcolor{green!10}\textbf{1.16} \\
    
    \rowcolor{green!10} \enspace + Distill ($n_{\mathrm{step}}=4$)
    & \textbf{48.96} & \textbf{30.95} & \cellcolor{blue!5} \underline{7.53}
    & \textbf{15.46} & \textbf{21.58}
    & \textbf{27.76} & \textbf{39.14}
    & \textbf{22.97} & \textbf{1.16} \\
    \bottomrule
    \bottomrule
    \end{tabular*}
    \raggedright
    \vspace{1mm}
    \scriptsize
    
    AUC and VUS are reported as percentages. Bold and underlined values indicate the best and second-best results, respectively.
    
    $^\dagger$RFMPose does not report mean errors on ROPE and does not release pretrained model weights, preventing evaluation under our protocol.
    
    %ComPose~\cite{ren2026compose} and Flow6D~\cite{mei2026flow6d} are excluded because neither releases public code or pretrained model weights.
\end{table*}

\begin{figure*}[!t]
    \centering
    \includegraphics[width=\textwidth]
    {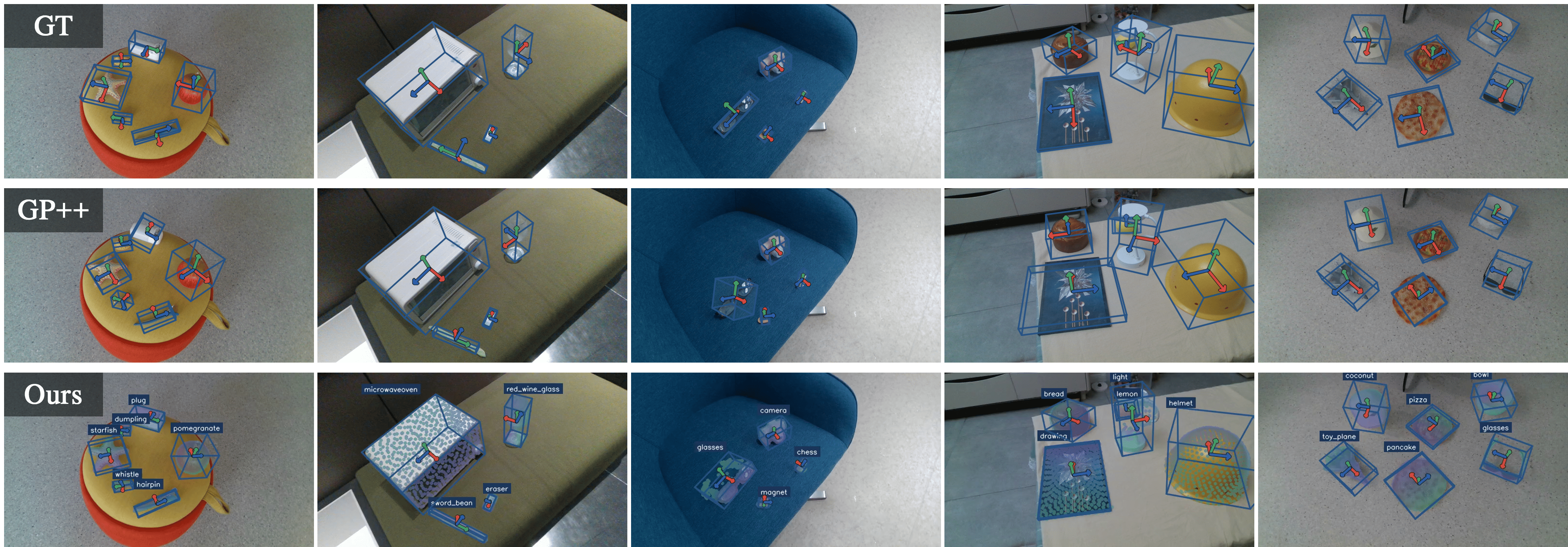}
    \caption{\textbf{Qualitative comparison on Omni6DPose ROPE}. From top to bottom: ground truth, GenPose++ (GP++), and ASM-6D. Predicted 3D bounding boxes and coordinate frames visualize the pose, while ASM-6D additionally recovers the fitted object shape from its generated correspondences.}
    \label{fig:asm6d_comparison}
\end{figure*}

\section{Evaluation}
\label{sec:evaluation}
\subsection{Experimental Setup}
\label{sec:experimental_setup}

\noindent\textbf{Implementation \& Hyperparameters.}
Unless otherwise specified, ASM-6D uses $N=642$ template points, a four-step distilled generator, $H=40$ correspondence hypotheses, 20 ADMM iterations, and visibility-weighted fitting-residual selection. We set $\lambda_{\mathbf c}=\sigma=10^{-5}$. These hyperparameter choices balance point-cloud density, inference speed, and convergence stability; they were tuned once on a held-out validation set of Omni6DPose~\cite{Zhang2024Omni6DPoseAB} and kept fixed across all subsequent benchmarks without task-specific tuning unless stated otherwise. All experiments are executed on a single NVIDIA RTX~3090~Ti GPU. Detailed architectural parameters and training configurations are provided in Appendix~\ref{appx:training_eval_details}.

\noindent\textbf{Datasets \& Training Protocols.}
We evaluate on two complementary category-level pose benchmarks:
\begin{itemize}[leftmargin=*, noitemsep, topsep=2pt]
  \item \textbf{Omni6DPose (SOPE $\to$ ROPE):} Models are trained strictly on the synthetic SOPE dataset~\cite{Zhang2024Omni6DPoseAB}, which contains 475K images generated from an ASM library of 4,385 CAD models across 156 categories. We evaluate zero-shot transfer on ROPE, comprising 65,454 real-world test images. Crucially, no ROPE images or scene instances are used during training or parameter tuning.
  \item \textbf{HouseCat6D:} Evaluated using its official 34/2/5 scene split (containing 16,884 training, 1,120 validation, and 3,360 test frames across 10 household categories). Models for HouseCat6D are trained entirely independently on its training split without combining or co-training with SOPE/ROPE data unless stated otherwise. Category priors are constructed solely from HouseCat6D training CAD meshes.
\end{itemize}
ROPE evaluates performance using a sensor-assisted global-depth refinement step, whereas HouseCat6D disables this refinement per its official evaluation protocol.

\noindent\textbf{Baselines.}
On ROPE, we compare against leading discriminative and generative pose estimators, including GenPose~\cite{zhang2024generative}, GenPose++~\cite{Zhang2024Omni6DPoseAB}, and RFMPose~\cite{ouyang2026rfmpose}. On HouseCat6D, we evaluate against recent state-of-the-art methods: AG-Pose~\cite{Lin_2024_CVPR}, SpotPose~\cite{Ren_2025_CVPR}, GCE-Pose~\cite{Li_2025_CVPR}, and ComPose~\cite{ren2026compose}. All results follow standard benchmark metrics under their respective protocols.

\subsection{Category-Level Pose Estimation on ROPE}
\label{sec:pose_estimation}

We evaluate category-level pose estimation on ROPE using 3D IoU AUC and joint rotation--translation volume under the surface (VUS). Table~\ref{tab:comparison_and_adaptation} compares ASM-6D with reported prior results and model variants. All ASM-6D rows use $H=40$ hypotheses; non-distilled models use ten integration steps, while distilled models use the indicated step count.

ASM-6D outperforms prior category-level pose methods, supporting generative correspondence prediction before geometric fitting. Encoder adaptation further improves accuracy, while few-step distillation preserves these gains with fewer decoder evaluations. Fig.~\ref{fig:asm6d_comparison} shows qualitative comparisons with GenPose++~\cite{Zhang2024Omni6DPoseAB}.

\begin{table}[!t]
\centering
\caption{Core design ablations on ROPE.}
\label{tab:core_ablations}
\scriptsize
\begin{tabular}{llcc}
\toprule
\toprule
Factor & Setting
& IoU$_{50}$ AUC $\uparrow$
& $10^\circ$/5\,cm VUS $\uparrow$ \\
\midrule

\multirow{2}{*}{Prediction}
& Regression ($H=1$)     & 11.58 & 16.98 \\
& \cellcolor{green!10}\textbf{Flow matching} ($H=1$)  & \cellcolor{green!10}\textbf{25.91} & \cellcolor{green!10}\textbf{29.24} \\
\midrule

\multirow{3}{*}{Hypotheses}
& $H=1$                  & 30.07 & 37.73 \\
& $H=4$                  & 30.59 & 38.65 \\
& \cellcolor{green!10}\textbf{$H=40$}                 & \cellcolor{green!10}\textbf{30.95} & \cellcolor{green!10}\textbf{39.14} \\
\midrule

\multirow{2}{*}{Shape model}
& Category mean          & 19.82 & 34.63 \\
& \cellcolor{green!10}\textbf{Full ASM}               & \cellcolor{green!10}\textbf{30.95} & \cellcolor{green!10}\textbf{39.14} \\
\bottomrule
\bottomrule
\end{tabular}
  \raggedright
  \vspace{1mm}
  \scriptsize
  
  Each block varies only the indicated factor. Full results are reported in Appendix~\ref{app:additional_ablations}.
\end{table}

\subsection{Core Ablations on ROPE}
\label{sec:core_ablations}
Table~\ref{tab:core_ablations} summarizes the principal design ablations. With the same frozen RGB-D encoder and $H=1$, flow-based prediction substantially outperforms deterministic regression, showing that its benefit does not arise solely from multi-hypothesis selection.
% \nfcomment{Could you provide FPS or ms for each of the factors in Table II?}

Increasing the number of hypotheses provides additional but diminishing gains, making $H$ primarily an accuracy--throughput tradeoff. Section~\ref{sec:shape-conditioned-pose} further shows that sampling preserves distinct shape--pose explanations under ambiguous observations. Active shape variation has a larger effect where fitting the full ASM clearly outperforms using only the category mean, while a small number of PCA modes captures most of the benefit. Together, these results show that generative prediction, hypothesis sampling, and instance-level shape variation each contribute to the final estimate. Full results are reported in Appendix~\ref{app:additional_ablations}.

% We separately evaluate the contributions of generative prediction, multiple hypotheses, and active shape variation. With the same frozen RGB-D encoder and $H=1$, flow-based prediction increases IoU$_{50}$ AUC from $11.58$ to $25.91$ and $10^\circ$/5\,cm VUS from $16.98$ to $29.24$ relative to deterministic regression. This comparison isolates the generative prediction procedure from
% multi-hypothesis selection.

% Using the four-step distilled model, increasing the hypothesis count from $H=1$ to $H=40$ raises IoU$_{50}$ AUC from $30.07$ to $30.95$ and $10^\circ$/5\,cm VUS from $37.73$ to $39.14$. Most of the gain is obtained with a small number of hypotheses: $H=4$ reaches $38.65$ VUS, only $0.49$ points below $H=40$, while increasing known-category throughput from $10.26$ to $30.24$ FPS.

% Active shape variation provides a substantially larger effect. With the generated hypotheses held fixed, fitting only the category mean achieves $19.82$ IoU$_{50}$ AUC and $34.63$ VUS. Allowing PCA shape variation increases these values to $30.95$ and $39.14$, respectively. Four retained modes already achieve $29.12$ IoU$_{50}$ AUC and $38.51$ VUS. These results show that category geometry, instance-level shape variation, and hypothesis sampling contribute separately to the final pose estimate. Full ablations are reported in Appendix~\ref{app:additional_ablations}.

\subsection{Category-Level Pose Estimation on HouseCat6D}
\label{sec:housecat6d}

We evaluate ASM-6D on HouseCat6D using the public AG-Pose GT-instance scoring protocol~\cite{Lin_2024_CVPR}. Both ASM-6D configurations use four-step distillation, $H=40$, and category ASMs constructed exclusively from HouseCat6D training CADs. The encoder receives unmasked completed depth, and sensor-assisted global-depth refinement is disabled. Training and supervision details are provided in Appendix~\ref{appx:housecat_training}.

% ASM-6D transfers effectively without additional pose pretraining, while SOPE initialization provides a further gain, showing that the learned correspondence representation transfers across datasets (Table~\ref{tab:housecat6d}). 
% \nfcomment{Do you actually have concrete results supporting this statement? i.e., do you have a ASM trained on one dataset and evaluated on other? Otherwise, not sure how you can justify this claim with the presented results so far.}
Table~\ref{tab:housecat6d} separates three training regimes. ASM-6D trained only on HouseCat6D already outperforms the prior methods across all reported metrics. Evaluating the SOPE-trained model directly on HouseCat6D without adaptation retains nontrivial accuracy, providing direct evidence that the learned correspondence representation transfers across datasets. Using the SOPE model to initialize HouseCat6D training further improves IoU$_{50}$ from $85.7$ to $90.7$ and $10^\circ$/5\,cm accuracy from $71.1$ to $79.7$. We report this setting separately because it uses additional pose-supervised pretraining.

\subsection{ASM Solver Analysis}
\label{sec:asm-solver-evaluation}
We evaluate the efficiency and numerical reliability of the ASM fitting
backend.

\begin{table}[t]
% \caption{Full-pipeline throughput in FPS for different hypothesis counts and
% ADMM iteration budgets. ``Known'' fits the known class prior, while ``All''
% fits all $156$ priors.}
\caption{Full-pipeline throughput across hypothesis and ADMM iteration budgets in FPS.}
\label{tab:full_pipeline_fps}
\centering
\setlength{\tabcolsep}{2.5pt}
\renewcommand{\arraystretch}{1.0}
\scriptsize
\resizebox{\columnwidth}{!}{%
\begin{tabular}{c|cccc|cccc}
    \hline
    \multirow{2}{*}{$H$}
    & \multicolumn{4}{c|}{Known single category}
    & \multicolumn{4}{c}{All $156$ priors} \\
    & 3 iter. & 5 iter. & 10 iter. & 20 iter.
    & 3 iter. & 5 iter. & 10 iter. & 20 iter. \\
    \hline
    1
    & 38.85 & 38.14 & 37.19 & 34.04
    & 28.66 & 28.37 & 27.05 & 25.11 \\
    4
    & 31.50 & 31.32 & 30.51 & 30.24
    & 22.93 & 23.03 & 22.04 & 20.65 \\
    8
    & 26.26 & 26.04 & 26.14 & 25.73
    & 18.97 & 18.88 & 18.33 & 17.24 \\
    16
    & 18.51 & 18.75 & 18.64 & 18.54
    & 13.30 & 13.14 & 12.88 & 12.43 \\
    32
    & 12.28 & 12.29 & 12.22 & 12.16
    & 8.43 & 8.34 & 8.26 & 8.02 \\
    40
    & 10.39 & 10.34 & 10.33 & 10.26
    & 7.13 & 7.08 & 6.98 & 6.82 \\
    \hline
\end{tabular}%
}
\end{table}

\begin{figure}[!t]
    \centering
    \includegraphics[width=\linewidth]
    {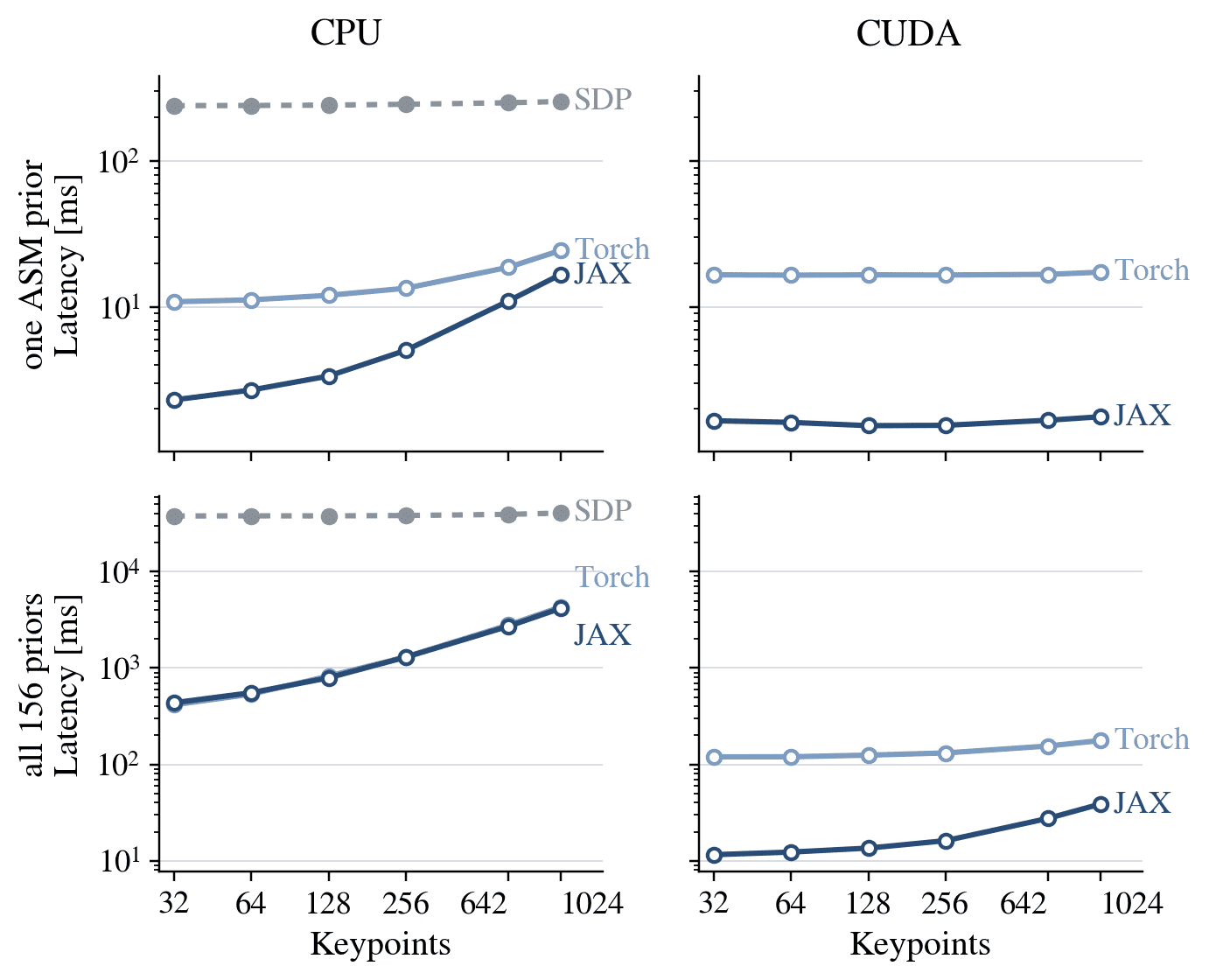}
    \caption{ASM solver latency versus keypoint count with $40$ hypotheses, $32$ PCA modes, and $20$ ADMM iterations. Columns compare CPU and CUDA, while rows compare one-prior fitting with exhaustive fitting over all $156$ priors. Timings include problem construction, optimization, scaled-rotation projection, and parameter reconstruction, but exclude correspondence inference and JIT compilation.}
    \label{fig:asm-solver-scaling}
\end{figure}

\noindent\textbf{Runtime.}
% Unknown-class inference fits each of the $H=40$ correspondence sets to all $C=156$ ASM priors, yielding $HC=6240$ independent reduced $9\times9$ problems per object. Fig.~\ref{fig:asm-solver-scaling} compares one-prior fitting with this exhaustive setting. At $N=642$, JAX/CUDA requires $1.68\,\mathrm{ms}$ for one prior and $27.51\,\mathrm{ms}$ for all $156$ priors. Although the number of fits increases by $156\times$, latency increases by only $16\times$ because the problems are processed in a single batch. For the exhaustive workload, JAX/CUDA is $5.6\times$ faster than PyTorch/CUDA ($155.09\,\mathrm{ms}$) and $1430\times$ faster than the serial float64 SDP reference ($39.39\,\mathrm{s}$). These measurements isolate the geometric backend; full-pipeline throughput is reported in Appendix~\ref{app:additional_ablations}.
Unknown-class inference requires fitting each $H=40$ correspondence hypothesis to all $C=156$ candidate ASMs. Fig.~\ref{fig:asm-solver-scaling} shows that batched JAX/CUDA scales much more favorably than serial or less optimized implementations, making exhaustive prior evaluation practical. Full-pipeline throughput is reported in Table~\ref{tab:full_pipeline_fps}. The tradeoff between speed and accuracy is detailed in Appendix~\ref{app:additional_ablations}.
% \nfcomment{Bring this table and corresponding paragraph here.}

\begin{table}[!t]
  \centering
  \caption{Category-level pose estimation on HouseCat6D.}
  \label{tab:housecat6d}

  \setlength{\tabcolsep}{1.7pt}
  \scriptsize
  \resizebox{\columnwidth}{!}{%
  \begin{tabular}{lccccccc}
  \toprule
  \toprule
  Method
  & Shape
  & IoU$_{25}$$\uparrow$
  & IoU$_{50}$$\uparrow$
  & $5^\circ$/2$\uparrow$
  & $5^\circ$/5$\uparrow$
  & $10^\circ$/2$\uparrow$
  & $10^\circ$/5$\uparrow$ \\
  \midrule

  AG-Pose~\cite{Lin_2024_CVPR}
  & -- & 88.1 & 76.9 & 21.3 & 22.1 & 51.3 & 54.3 \\
  SpotPose~\cite{Ren_2025_CVPR}
  & -- & 89.1 & 77.0 & 23.8 & 24.5 & 52.3 & 54.8 \\
  GCE-Pose~\cite{Li_2025_CVPR}
  & \checkmark & -- & 79.2 & 24.8 & 25.7 & 55.4 & 58.4 \\
  ComPose~\cite{ren2026compose}
  & \checkmark & 90.3 & 80.6 & 25.8 & 27.6 & 57.8 & 61.5 \\

  \rowcolor{green!10} \textbf{ASM-6D (HC6D)}
  & \checkmark
  & \textbf{91.0}
  & \textbf{85.7}
  & \textbf{30.2}
  & \textbf{32.7}
  & \textbf{64.3}
  & \textbf{71.1} \\

  \midrule
  \multicolumn{8}{l}{\textit{Additional SOPE pose-task pretraining}} \\

  ASM-6D (SOPE)
  & \checkmark & 56.6 & 33.9 & 9.6 & 10.8 & 24.5 & 27.9 \\

  \rowcolor{green!10} \textbf{ASM-6D (SOPE$\rightarrow$HC6D)}$^\dagger$
  & \checkmark & \textbf{95.5} & \textbf{90.7} & \textbf{40.9} & \textbf{42.8} & \textbf{74.2} & \textbf{79.7} \\

  \bottomrule
  \bottomrule
  \end{tabular}%
  }
  \raggedright
  \vspace{1mm}
  \scriptsize
  
  Both ASM-6D variants use four-step distillation, $H=40$, and unmasked completed depth under the public AG-Pose GT-instance scoring protocol.
  
  $^\dagger$Additional pose-supervised pretraining on SOPE before HouseCat6D adaptation.
\end{table}

\noindent\textbf{Global Optimality Certificate.}
We validate the reduced fits against the SDP relaxation using the numerical certificate described in Appendix~\ref{appx:sdp_kkt}. 
% Across deterministic tests spanning noise, partial visibility, and outliers, all evaluated solutions satisfy the certificate conditions after the standard ADMM budget. This indicates that the reduced problems are solved to globally optimal solutions in the tested regimes, while not implying universal tightness for arbitrary model outputs.
% \nfcomment{Where are these tests? Where is the evidence? They should at least be described/summarized in the appendix if you are mentioning them here.}
We construct the SDP relaxation from the same float64 normal equations and solve it with Clarabel, using SCS as a fallback. With $\epsilon=10^{-4}$, a candidate passes the numerical certificate when its scaled-rotation violation is at most $\epsilon$, the minimum dual-slack eigenvalue is at least $-\epsilon$, the stationarity residual is at most $10\epsilon$, and its objective is no more than $10\epsilon$ below the SDP lower bound.

All $16$ deterministic cases spanning clean measurements, noise, partial visibility, and outliers satisfy these conditions after $20$ float32 JAX updates for all 4,385 CAD models from the SOPE dataset~\cite{Zhang2024Omni6DPoseAB}. The median relative SDP gap is $1.5\times10^{-7}$, the worst gap is $3.8\times10^{-6}$, and the maximum rank ratio is $1.8\times10^{-6}$. These results certify the tested reduced problems.

\begin{table*}[t]
    \centering
    \caption{\textbf{Shape-completion results.} \textsc{M-PC}: instance-mask segmented depth point cloud; \textsc{ROI}: RGB-D crop and box.}
    \label{tab:shape_completion_main}
    \small
    \setlength{\tabcolsep}{3.0pt}
    \renewcommand{\arraystretch}{1.06}
    \begin{tabular}{lll|ccc|ccc}
        \toprule
        \toprule
        & & & \multicolumn{3}{c|}{SOPE test, stride 3 ($n=339{,}840$)} & \multicolumn{3}{c}{RealNOCS \texttt{real\_test} ($n=14{,}077$)} \\
        Method & Input & Prior & F1@1\%$\uparrow$ & CD-L1(m)$\downarrow$ & HD95(m)$\downarrow$ & F1@1\%$\uparrow$ & CD-L1(m)$\downarrow$ & HD95(m)$\downarrow$ \\
        \midrule
        PoinTr~\cite{yu2021pointr} & \textsc{M-PC} & -- & 0.0521 & 0.1181 & 0.2416 & 0.0277 & 0.0452 & 0.0809 \\
        SnowflakeNet~\cite{xiang2021snowflakenet} & \textsc{M-PC} & -- & 0.0498 & 0.0782 & 0.1721 & 0.0215 & 0.0509 & 0.1014 \\
        AdaPoinTr~\cite{yu2023adapointrdiversepointcloud} & \textsc{M-PC} & -- & 0.0445 & 0.1045 & 0.1763 & 0.0146 & 0.0543 & 0.0867 \\
        DR-Pose$^\dagger$~\cite{zhou2023dr} & \textsc{M-PC} & mean & -- & -- & -- & \cellcolor{green!10}\textbf{0.0524} & \cellcolor{blue!5}\underline{0.0195} & 0.0232 \\
        \midrule
        \rowcolor{green!10} ASM-6D w/o prior & \textsc{ROI} & -- & \textbf{0.1749} & \textbf{0.0114} & \textbf{0.0146} & \cellcolor{blue!5}\underline{0.0516} & \textbf{0.0185} & \textbf{0.0199} \\
        ASM-6D w/ prior & \textsc{ROI} & ASM & \cellcolor{blue!5} \underline{0.1647} & \cellcolor{blue!5} \underline{0.0122} & \cellcolor{blue!5} \underline{0.0163} & 0.0473 & 0.0198 & \cellcolor{blue!5}\underline{0.0209} \\
        \bottomrule
        \bottomrule
    \end{tabular}
    
    \vspace{0.5mm}
    % \raggedright
    \centering
    \footnotesize
    $^\dagger$DR-Pose is trained on RealNOCS train and is reported only on RealNOCS. Bold and underline denote best and second-best.
\end{table*}

\section{Representation Analysis and Validation of Downstream Applications}
\label{sec:reuse}

The preceding experiments evaluated the performance of ASM-6D for pose and shape estimation against SOTA baselines on a diverse set of large benchmark datasets. Apart from showing that ASM-6D exceeds performance on all benchmarks due to our combination of robust and complete keypoint correspondence mapping and an ADMM solver to estimate the shape-pose parameters, we now evaluate our representation on geometry-aware tasks such as shape-completion, dealing with pose ambiguities (a critical issue for symmetric objects), prior selection analysis, and real-time object tracking and grasping.
% We next examine whether the same representation supports additional geometric tasks without changing its underlying indexed structure.

\begin{figure*}[!t]
    \centering
    \includegraphics[width=\textwidth]
    {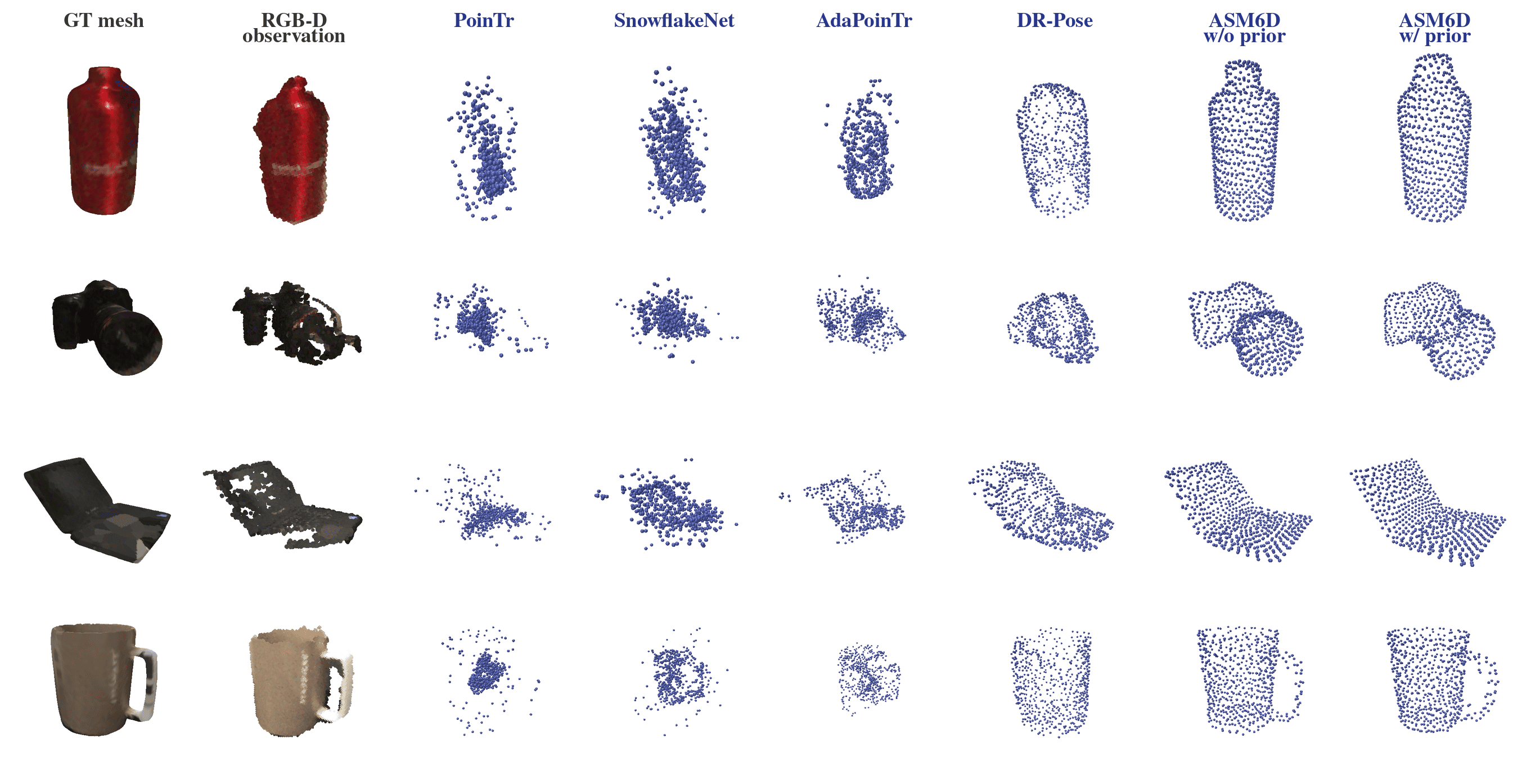}
    \caption{\textbf{Full-pipeline RealNOCS completion at the native $642$-point metric budget.} All outputs are shown in a common ground-truth camera frame and scene scale. Point-cloud methods use the released Mask R-CNN partial cloud, whereas ASM-6D uses the object ROI without mask pixels. RGB-D colors are used only for visualization. 
    % \nfcomment{This figure is not referenced anywhere in the main text. }
    }
    \label{fig:shape_completion_realnocs}
\end{figure*}

\begin{figure}[!t]
    \centering
    \includegraphics[width=\columnwidth]
    {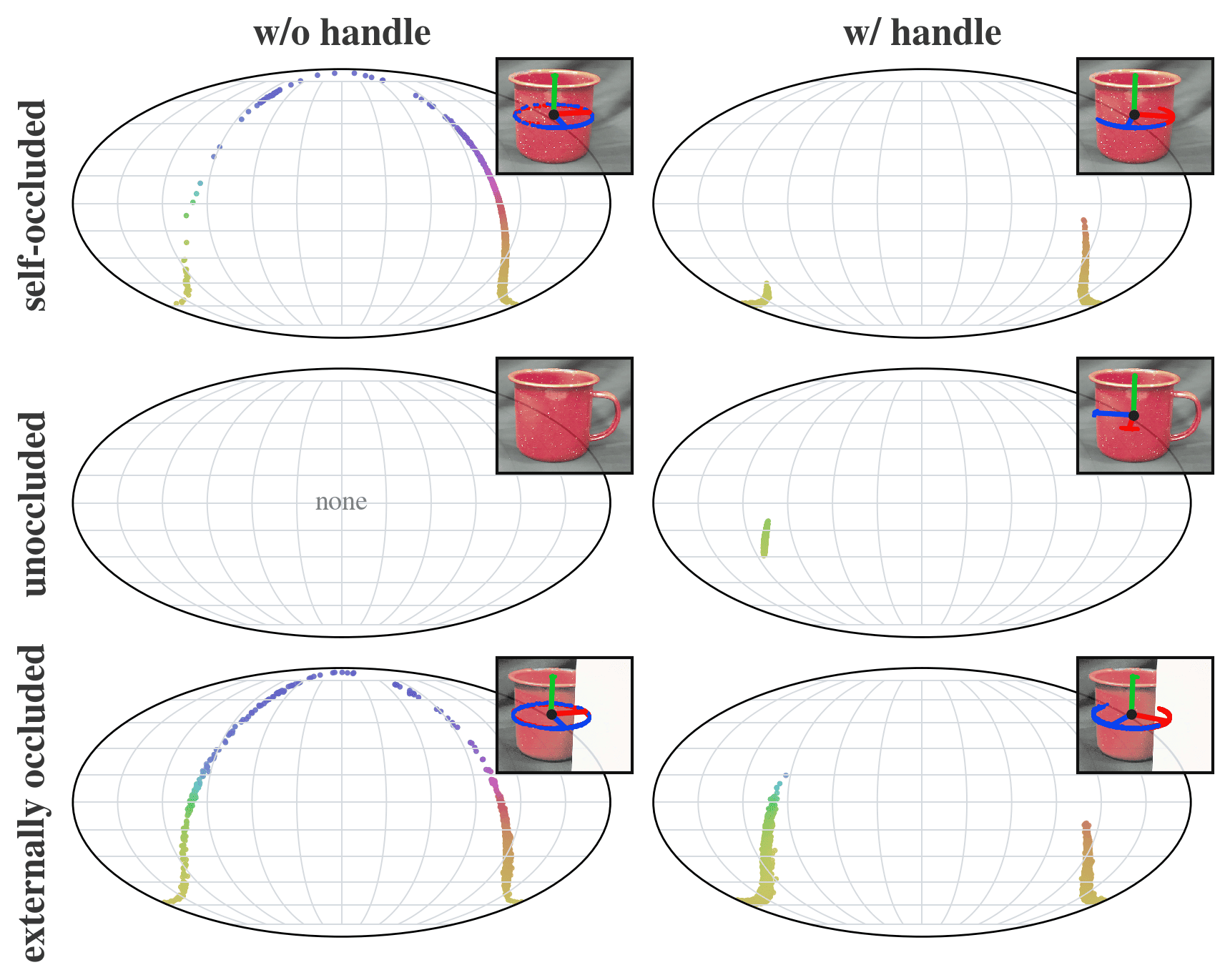}
    \caption{\textbf{Shape-conditioned pose ambiguity for one cup under three visibility conditions}. Each point represents an independently sampled correspondence set after ASM fitting. Mollweide position and color encode rotation, while columns separate geometry-derived handle-free and handle-bearing completions.}
    \label{fig:shape-conditioned-pose}
\end{figure}

\subsection{Shape Completion}
\label{sec:shape_completion}
We evaluate the raw correspondence hypotheses as complete geometry on SOPE and RealNOCS using F1@1\%, CD-L1, and HD95. Point-cloud baselines receive instance-mask-segmented depth points, whereas ASM-6D receives an RGB-D object ROI. Table~\ref{tab:shape_completion_main} is therefore a native-interface comparison rather than an input-matched comparison. Full evaluation details are provided in Appendix~\ref{appx:completion_protocol}.

ASM-6D performs strongly as a completion model despite representing shape as indexed correspondences rather than an unordered point cloud as in Fig.~\ref{fig:shape_completion_realnocs}. This supports a key property of the representation. Each sample contains both completed geometry and persistent point identities. ASM fitting slightly reduces pointwise completion accuracy because it projects the prediction onto a lower-dimensional category shape space, but in return provides the structured pose, metric scale, and shape representation used downstream.

\subsection{Shape-Conditioned Pose Ambiguity}
\label{sec:shape-conditioned-pose}

A partial observation can support multiple shape completions with different pose ambiguities. We study one cup under self-occlusion, no occlusion, and external occlusion. For each capture, we sample $16{,}384$ correspondence sets and fit the mug ASM independently. A geometry-based radial test separates approximately handle-free and handle-bearing completions.

Fig.~\ref{fig:shape-conditioned-pose} shows that a visible handle yields a single concentrated pose mode, while occlusion preserves multiple completion types with distinct azimuthal distributions. A linear classifier using only the fitted ASM coefficients separates these groups with cross-validated AUCs of $0.983$ and $0.997$ under self- and external occlusion, respectively. Thus, correspondence sampling preserves coupled shape--pose explanations of the same partial observation.

\subsection{Geometry-Based Prior Selection}
\label{sec:classification}

We next test whether the sampled correspondences retain enough category-level geometry to select a shape prior without an input class label. The same $H=40$ correspondence sets are fitted to all $156$ ASMs, and ranking uses only the geometric fitting residual. The selector reaches $73.7\%$ on RealNOCS and $76.5\%$ on SOPE, far above the $0.64\%$ chance rate for $156$ classes. Distinctive categories such as bowls, laptops, and mugs are selected reliably, while most errors occur among geometrically similar categories such as bottles and cans. 

\begin{figure}[t]
\centering
\includegraphics[width=0.49\columnwidth]
{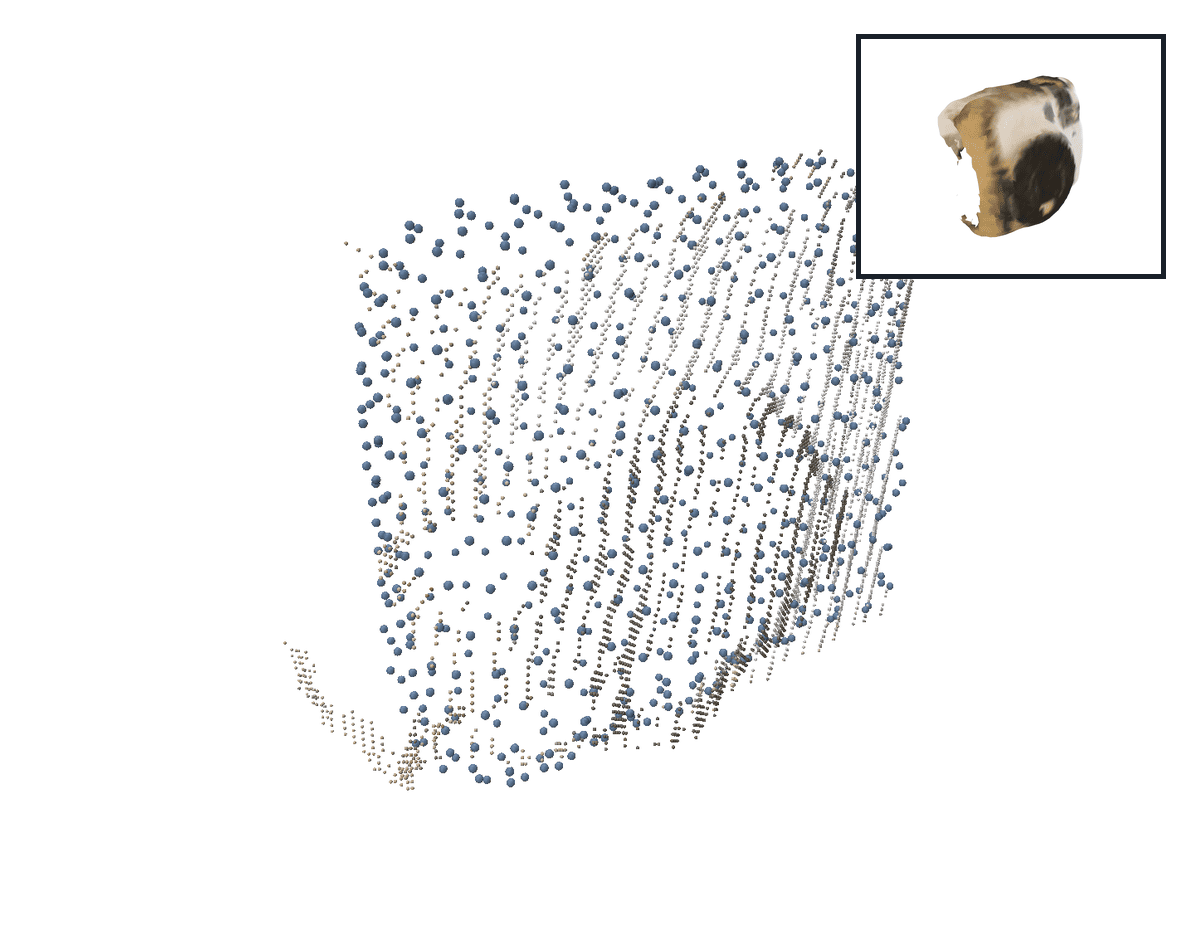}
\hfill
\includegraphics[width=0.49\columnwidth]
{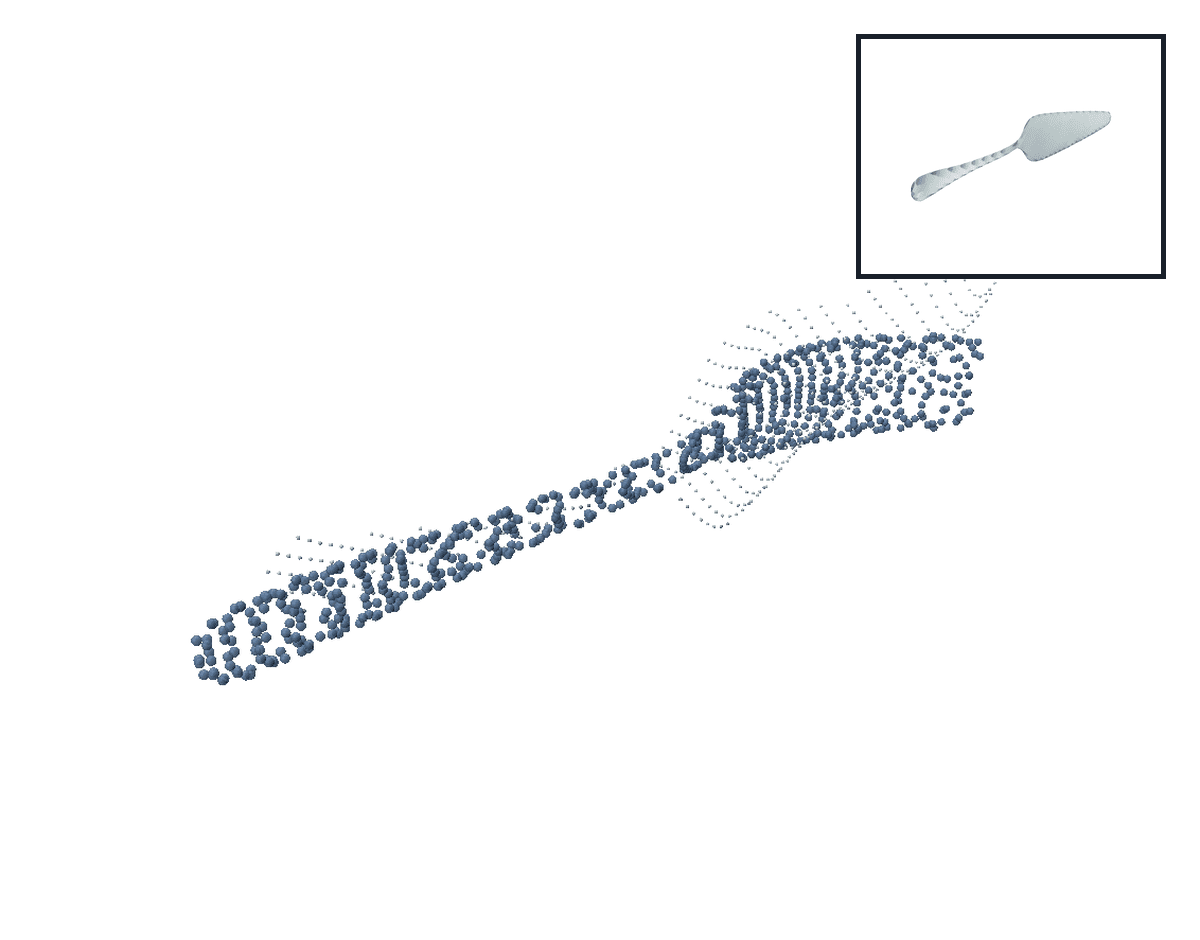}
\caption{\textbf{Representative exact-class errors.} Left: RealNOCS camera $\rightarrow$ handbag. Right: SOPE knife $\rightarrow$ fork. Each view shows the isolated RGB-D observation and selected incorrect ASM fit; the inset shows the ground-truth mesh for context.}
\label{fig:asm-class-confusions}
\end{figure}

As Fig.~\ref{fig:asm-class-confusions} shows, even an incorrect prior can still provide a plausible geometric fit. These results indicate that the sampled correspondences retain strong category-level geometric information without a separate classifier.

\begin{table}[t]
\centering
\caption{Exact canonical-class selection through exhaustive ASM fitting.}
\label{tab:asm-classification}
\small
\begin{tabular}{lcc}
\toprule
\toprule
Dataset / class & Top-1 & Correct / total \\
\midrule
RealNOCS (all) & 73.7\% & 11884/16117 \\
Bottle         & 47.0\% & 1295/2757 \\
Bowl           & 89.0\% & 2109/2370 \\
Camera         & 64.8\% & 1729/2669 \\
Can            & 43.0\% & 1174/2728 \\
Laptop         & 100.0\% & 2754/2754 \\
Mug            & 99.4\% & 2823/2839 \\
\midrule
\rowcolor{green!10} \textbf{SOPE (all)}     & \textbf{76.5\%} & \textbf{4614/6035} \\
\bottomrule
\bottomrule
\end{tabular}
\end{table}

\begin{figure*}[!t]
    \centering
    \includegraphics[width=\textwidth]
    {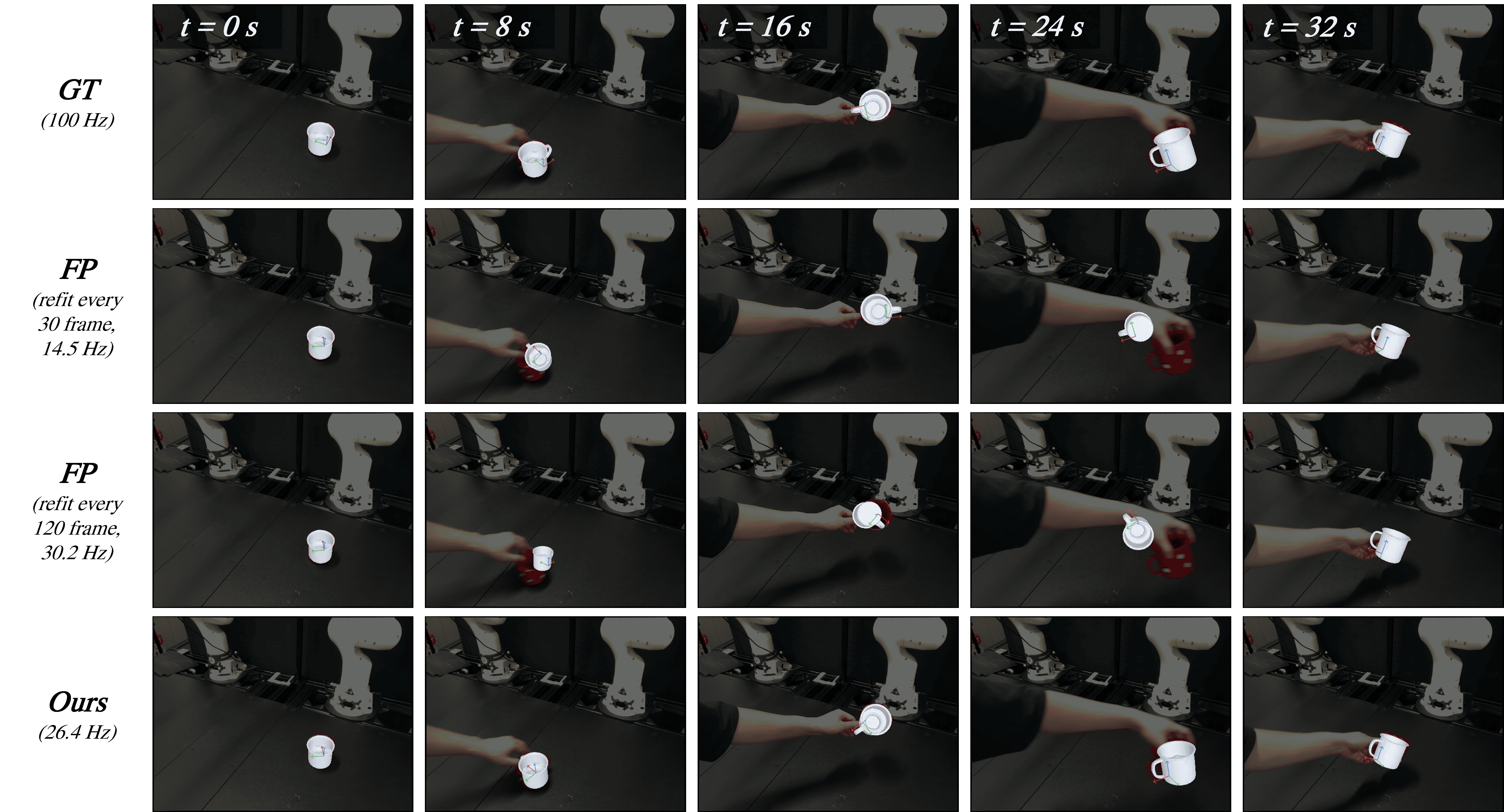}
    % \caption{Tracking on the high-motion mug sequence. ASM-6D maintains a stable pose by selecting among newly sampled correspondence sets, whereas native FoundationPose accumulates substantial drift under rapid translation and rotation.}
    \caption{\textbf{Tracking on the high-motion mug sequence.} Rows show ground truth, FoundationPose (FP) with periodic re-registration, and ASM-6D over time. The comparison illustrates the temporal behavior of the evaluated tracking configurations under rapid object motion (mean translation speed: $0.197~\mathrm{m\,s^{-1}}$, mean rotation speed: $88.2^\circ\,\mathrm{s^{-1}}$).
    % \nfcomment{Add average speed of the mug and computation time/rate of ASM-6D and the other methods. Since GT is Mocap, defined it in the figure and add the HZ, same for foundation pose variants.}
    }
    \label{fig:tracking_comparison}
\end{figure*}
\begin{figure*}[!t]
    \centering
    \includegraphics[width=\textwidth]
    {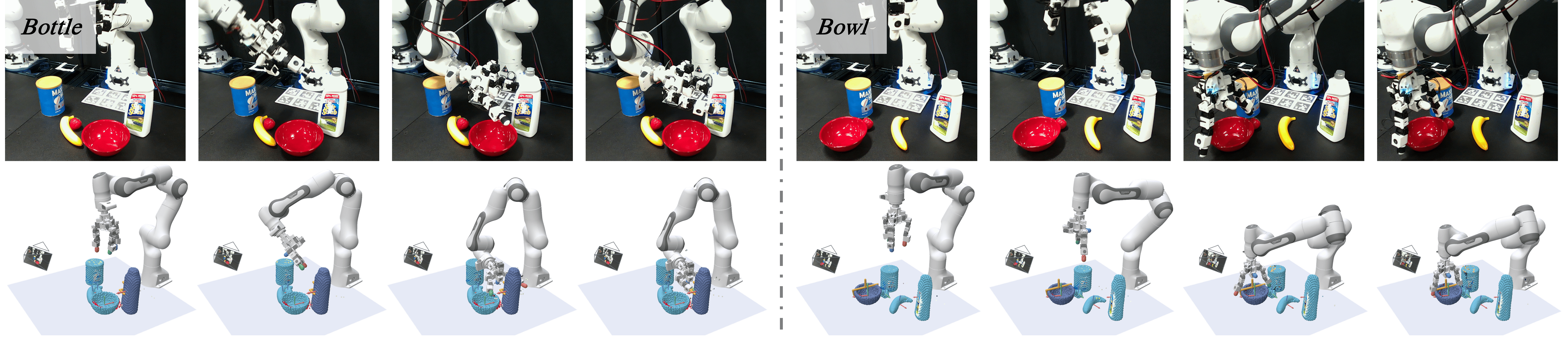}
    \caption{\textbf{Qualitative object grasping experiments.} In the tabletop scene, ASM-6D estimates the pose and shape of target objects among multiple surrounding objects, enabling grasping of the bottle and bowl.}
    \label{fig:grasp}
\end{figure*}

\begin{table}[!t]
  \centering
  \caption{Tracking accuracy on HO3D, pooled over scored frames.}
  \label{tab:tracking-ho3d}
  \footnotesize
  \setlength{\tabcolsep}{3pt}
  \begin{tabular}{lrrrrr}
    \toprule
    \toprule
    Method
    & \multicolumn{2}{c}{Trans. (cm)$\downarrow$}
    & \multicolumn{2}{c}{Rot. ($^\circ$)$\downarrow$}
    & ADD (cm)$\downarrow$ \\
    \cmidrule(lr){2-3}
    \cmidrule(lr){4-5}
    & Med. & P90 & Med. & P90 & Med. \\
    \midrule
   \textbf{ASM-6D}
    & 0.46 & 2.60 & 4.2 & 53.9 & 0.57 \\
    \rowcolor{green!10} FP (native)
    & 0.11 & 0.24 & 1.9 & 7.2 & 0.20 \\
    FP (30f reinit.)
    & 0.13 & 1.16 & 2.7 & 179.6 & 0.23 \\
    \bottomrule
  \end{tabular}
\end{table}

\subsection{Tracking}
\label{sec:tracking}
We evaluate temporal use of the correspondence representation in two tracking settings. ASM-6D estimates pose, scale, and shape on the first frame, retains scale and shape over the sequence, and generates new correspondence hypotheses for subsequent pose updates. Translation and rotation estimates are smoothed using One-Euro filters~\cite{casiez20121}. FoundationPose~\cite{wen2024foundationpose} receives the exact YCB instance CAD, whereas ASM-6D uses a category-level ASM. Detailed tracker configurations are provided in Appendix~\ref{appx:tracking_protocol}.

On seven HO3D~\cite{hampali2020honnotate} sequences comprising 448 scored frames, native FoundationPose (FP) obtains the lowest aggregate errors. ASM-6D achieves median translation, rotation, and ADD errors of $0.46$\,cm, $4.2^\circ$, and $0.57$\,cm, respectively, using only the category-level shape model (Table~\ref{tab:tracking-ho3d}). Thus, HO3D primarily demonstrates that the sampled correspondence representation can be reused effectively over time, while an exact-CAD tracker remains advantageous under relatively favorable conditions with modest motion and limited occlusion.
% \nfcomment{What do you want to convey with these results? It seems FP(native) works quite well. Why? are there any occlusions here? Is it slow motion? Provide some insight on these results. Are there any frames/pictures of these result?}

\begin{table}[t]
  \centering
  \caption{Tracking accuracy and throughput on the high-motion mug sequence. $^*$FP is FoundationPose~\cite{wen2024foundationpose}}
  \label{tab:tracking-dynamic}
  \footnotesize
  \setlength{\tabcolsep}{3pt}
  \begin{tabular}{lrrrrrr}
    \toprule
    Method
    & \multicolumn{2}{c}{Trans. (cm)$\downarrow$}
    & \multicolumn{2}{c}{Rot. ($^\circ$)$\downarrow$}
    & ADD (cm)$\downarrow$
    & FPS $\uparrow$\\
    \cmidrule(lr){2-3}
    \cmidrule(lr){4-5}
    & Med. & P90
    & Med. & P90
    & Med.
    & Steady / Amort. \\
    \midrule
    \rowcolor{green!10} \textbf{ASM-6D}
    & \textbf{1.45} & \textbf{2.74} & \textbf{12.3} & \textbf{35.4} & \textbf{1.56} & \textbf{28.3 / 26.4} \\
    FP (native)
    & 63.60 & 83.80 & 144.4 & 165.2 & 62.12 & 64.0 / 55.3 \\
    FP (30f reinit.)
    & 3.03 & 7.80 & 64.8 & 163.7 & 4.29 & 55.3 / 14.5 \\
    FP (60f reinit.)
    & 3.06 & 11.14 & 72.8 & 163.1 & 4.47 & 55.0 / 22.5 \\
    FP (120f reinit.)
    & 3.30 & 40.00 & 48.7 & 161.4 & 4.16 & 55.1 / 30.2 \\
    \bottomrule
    \bottomrule
  \end{tabular}
\end{table}

However, on the high-motion mug sequence, with mean translation and rotation speeds of $0.197~\mathrm{m\,s^{-1}}$ and $88.2^\circ\,\mathrm{s^{-1}}$, respectively, ASM-6D achieves median translation, rotation, and ADD errors of $1.45$\,cm, $12.3^\circ$, and $1.56$\,cm at $26.4$ amortized FPS (Table~\ref{tab:tracking-dynamic}, Fig.~\ref{fig:tracking_comparison}). Periodic FoundationPose re-registration reduces the large error of native tracking but does not match ASM-6D's reported aggregate accuracy on this sequence.
% Because the two systems differ in shape priors, registration, hypothesis generation, and temporal processing, these comparisons demonstrate temporal use of the representation rather than isolating a universal tracking advantage.
Together, these results suggest that exact-CAD tracking is highly effective under smooth motion and limited occlusion, whereas repeated hypothesis generation becomes more useful as motion and occlusion increase tracking ambiguity.

\subsection{Object Grasping}
Finally, we provide a qualitative grasping demonstration using the recovered metric pose and shape. Fig.~\ref{fig:grasp} shows tabletop scenes containing multiple objects, where ASM-6D estimates the target geometry and provides the pose and shape used to generate the robot grasp. The fitted ASM is transformed into the robot frame using the estimated Sim(3). A grasp generated from the fitted surface is then executed by biRRT~\cite{kuffner2000rrt}.
The bottle and bowl examples show that the recovered pose, scale, and shape can be used directly by a downstream grasping pipeline.
% \nfcomment{Give a short description of the grasping pipeline you used and how the recovered pose and shape were used. You can cite other works to make it brief, but it should be explicit how we use ASM-6D with a graping controller.}

% \section{Limitations and Future Work}
% ASM-6D requires an object crop and known camera intrinsics, but no pixel-wise mask. Integrating detection and tracking could reduce external preprocessing and reuse fitted states across frames.

% Unknown-class inference currently evaluates all 156 offline ASM priors. Larger libraries could use hierarchical retrieval, while automated shape standardization could simplify adding new categories.

% The current candidate selector relies on a simple visibility-weighted ASM fitting residual. While adding fitted-depth consistency provides no consistent improvement, more expressive ranking strategies could incorporate visibility, sensor depth, geometric fit structure, and image features while preserving the explicit ASM-fitting backend.

% Finally, posterior certification applies to the reduced geometric fit for a given hypothesis--prior pair, rather than to the learned correspondence model. Future work could use geometric consistency more directly during training while retaining the explicit and verifiable fitting stage.

\section{Conclusion}
% We presented ASM-6D, which represents category-level uncertainty with generative indexed correspondences. Conditional flow matching samples complete correspondence sets, including occluded points, which are fitted to ASMs to recover pose, metric scale, and shape. Batched ADMM efficiently fits many sampled correspondence sets to candidate shape priors, with posterior optimality certification when its conditions are satisfied.

% Experiments show improved category-level pose estimation and demonstrate reuse of the same correspondence sets for completion, ambiguity analysis, prior selection, tracking, and grasping. With $H=40$ hypotheses and $20$ ADMM iterations, the full pipeline runs at $10.3$ FPS for known-category inference and $6.8$ FPS for exhaustive $156$-prior selection. Overall, ASM-6D demonstrates that generative correspondences provide a practical interface between learned uncertainty and explicit geometric reasoning.

We presented ASM-6D, a category-level pose--shape estimation framework that generates alternative complete geometric interpretations directly in the correspondence space of a category-level shape model. Because the generated and parametric representations share point identities, each hypothesis can be fitted directly for pose, isotropic metric scale, and instance shape. Analytic elimination of translation and shape coefficients reduces each fit to a compact nine-variable scaled-rotation problem, enabling efficient batched evaluation of multiple hypotheses and candidate shape priors.

ASM-6D achieves state-of-the-art category-level 6D pose performance on both ROPE and HouseCat6D. Controlled ablations isolate the contributions of generative prediction, hypothesis sampling, and active shape variation. Additional analyses show that the same indexed hypotheses retain useful complete geometry, preserve distinct pose–shape explanations under occlusion, support finite-library prior selection, and can be reused for temporal tracking and downstream manipulation. Together, these results establish indexed correspondence hypotheses as a structured interface between learned geometric prediction and explicit pose--shape inference.

ASM-6D currently assumes a provided object ROI, known camera intrinsics, and an offline library of category-specific ASMs. Scaling to substantially larger shape libraries may require coarse-to-fine or learned prior retrieval rather than exhaustive fitting. Temporal inference could also be extended beyond the current filtering scheme by propagating pose--shape hypotheses with explicit motion or physics models. Finally, combining the indexed ASM representation with richer surface models, such as Gaussian splats~\cite{kerbl20233d}, could provide denser geometry for downstream tasks including dynamic manipulation, active perception, and grasp planning.

\ifCLASSOPTIONcaptionsoff
  \newpage
\fi



\bibliographystyle{IEEEtran}


% \clearpage
\appendices

\section{Proof of Proposition \ref{prop:qp}}
% \section{Closed-form Optimal Shape and Translation}
\label{appx:closed_form}

To prove the reduction from Eq.~\eqref{eq:admm_full} to Eq.~\eqref{eq:qcqp}, we first derive the optimal translation $\boldsymbol{\tau}^*$ and shape coefficients $\mathbf{c}^*$ for fixed $\mathbf{\Omega}$. The objective is:
\begin{equation}
    f(\mathbf{\Omega}, \mathbf{c}, \boldsymbol{\tau}) = \frac{1}{N} \sum_{i=1}^N w_i \left\| \mathbf{\Omega}\hat{\mathbf{y}}_i - (\mathbf{b}_i^0 + \mathbf{B}_i\mathbf{c}) - \boldsymbol{\tau} \right\|^2 + \lambda_\mathbf{c} \|\mathbf{c}\|^2
\end{equation}

\subsection{Derivation of Optimal Translation}
The optimal translation $\boldsymbol{\tau}^*$ is obtained by setting the partial derivative of $f$ with respect to $\boldsymbol{\tau}$ to zero:
\begin{equation}
    \nabla_{\boldsymbol{\tau}} f \big|_{\boldsymbol{\tau}=\boldsymbol{\tau}^*} = \frac{2}{N} \sum_{i=1}^N w_i \left( \boldsymbol{\tau}^* - \left( \mathbf{\Omega}\hat{\mathbf{y}}_i - \mathbf{b}_i^0 - \mathbf{B}_i\mathbf{c} \right) \right) = \mathbf{0}_{3\times1}
\end{equation}
Solving for $\boldsymbol{\tau}^*$ yields the weighted mean-based estimator:
\begin{equation}
    \boldsymbol{\tau}^* = \mathbf{\Omega} \bar{\mathbf{y}} - \bar{\mathbf{b}}^0 - \bar{\mathbf{B}} \mathbf{c}
\end{equation}
where the weighted barycentric coordinates for the observations, mean shape, and basis matrix are defined respectively as:
\begin{equation}
    \bar{\mathbf{y}} = \frac{\sum_{i=1}^N w_i \hat{\mathbf{y}}_i}{\sum_{i=1}^N w_i}, \quad 
    \bar{\mathbf{b}}^0 = \frac{\sum_{i=1}^N w_i \mathbf{b}_i^0}{\sum_{i=1}^N w_i}, \quad 
    \bar{\mathbf{B}} = \frac{\sum_{i=1}^N w_i \mathbf{B}_i}{\sum_{i=1}^N w_i}
\end{equation}

\subsection{Centering the Optimization}
Substituting $\boldsymbol{\tau}^*$ back into the original objective function allows us to simplify the expression using centered coordinates relative to the weighted means:
\begin{equation}
    \tilde{\mathbf{y}}_i = \hat{\mathbf{y}}_i - \bar{\mathbf{y}}, \quad 
    \tilde{\mathbf{b}}_i^0 = \mathbf{b}_i^0 - \bar{\mathbf{b}}^0, \quad 
    \tilde{\mathbf{B}}_i = \mathbf{B}_i - \bar{\mathbf{B}}
\end{equation}
This centering removes the explicit dependence on $\boldsymbol{\tau}$. We denote the resulting objective function as $f_{\boldsymbol{\tau}^*}$:
\begin{equation}
     f_{\boldsymbol{\tau}^*}(\mathbf{\Omega}, \mathbf{c}) = \frac{1}{N}\sum_{i=1}^N w_i \left\|\mathbf{\Omega} \tilde{\mathbf{y}}_i - (\tilde{\mathbf{b}}_i^0+\tilde{\mathbf{B}}_i\mathbf{c})\right\|^2 + \lambda_\mathbf{c} \|\mathbf{c}\|^2 
\end{equation}

\subsection{Rotation Linearization}
To make the formulation compact, we define the row-wise vectorization of the scaled-rotation matrix $\mathbf{\Omega}$. Let $\mathbf{v}_{i}^\top \in \mathbb{R}^3$ represent the $i$-th row of $\mathbf{\Omega}$:
\begin{equation}
    \mathbf{\Omega} = \begin{bmatrix} \mathbf{v}_{1}^\top \\ \mathbf{v}_{2}^\top \\ \mathbf{v}_{3}^\top \end{bmatrix}, \quad \boldsymbol{\omega} = \mathrm{vec}(\mathbf{\Omega}) = \begin{bmatrix} \mathbf{v}_{1} \\ \mathbf{v}_{2} \\ \mathbf{v}_{3} \end{bmatrix} \in \mathbb{R}^9
\end{equation}
We introduce the expanded observation matrix $\tilde{\mathbf{Y}}_i \in \mathbb{R}^{3\times 9}$ to satisfy the identity $\mathbf{\Omega}\tilde{\mathbf{y}}_i = \tilde{\mathbf{Y}}_i \boldsymbol{\omega}$:
\begin{equation}
    \tilde{\mathbf{Y}}_i = \mathbf{I}_3 \otimes \tilde{\mathbf{y}}_i^\top = 
    \begin{bmatrix}
         \tilde{\mathbf{y}}_i^\top & \mathbf{0} & \mathbf{0} \\
         \mathbf{0} & \tilde{\mathbf{y}}_i^\top & \mathbf{0} \\
         \mathbf{0} & \mathbf{0} & \tilde{\mathbf{y}}_i^\top   
    \end{bmatrix}
\end{equation}

\subsection{Optimal Shape Coefficients}
Setting the gradient $\nabla_{\mathbf{c}} f_{\boldsymbol{\tau}^*} = \mathbf{0}$ leads to the optimal shape coefficients as a linear mapping of the orientation vector:
\begin{equation}
    \mathbf{c}^* = \mathbf{M}_c \boldsymbol{\omega} + \mathbf{d}_c
\end{equation}
Using the regularized basis covariance matrix $\mathbf{\Gamma} = \sum_{i=1}^N w_i \tilde{\mathbf{B}}_i^\top \tilde{\mathbf{B}}_i + N\lambda_\mathbf{c} \mathbf{I}$, the mapping parameters are defined as:
\begin{equation}
    \mathbf{M}_c = \mathbf{\Gamma}^{-1} \left( \sum_{i=1}^N w_i \tilde{\mathbf{B}}_i^\top \tilde{\mathbf{Y}}_i \right), \quad 
    \mathbf{d}_c = -\mathbf{\Gamma}^{-1} \left( \sum_{i=1}^N w_i \tilde{\mathbf{B}}_i^\top \tilde{\mathbf{b}}_i^0 \right)
\end{equation}

\subsection{Quadratic Reduction}
Substituting the optimal coefficients $\mathbf{c}^*$ back into $f_{\boldsymbol{\tau}^*}$ reduces the objective to a function solely of the orientation vector $\boldsymbol{\omega}$. By grouping terms as:
\begin{equation}
    \tilde{\mathbf{M}}_i = \tilde{\mathbf{Y}}_i - \tilde{\mathbf{B}}_i \mathbf{M}_c, \quad \tilde{\mathbf{d}}_i = -(\tilde{\mathbf{B}}_i \mathbf{d}_c + \tilde{\mathbf{b}}_i^0)
\end{equation}
The objective can be rewritten in the standard quadratic form:
\begin{equation}
    f(\boldsymbol{\omega}) = \frac{1}{2}\boldsymbol{\omega}^\top \mathbf{H} \boldsymbol{\omega} + \mathbf{g}^\top \boldsymbol{\omega} + \text{const}
\end{equation}
The Hessian matrix $\mathbf{H} \in \mathbb{R}^{9 \times 9}$ and the linear gradient vector $\mathbf{g} \in \mathbb{R}^9$ are:
\begin{equation}
\label{eq:qp-terms}
    \begin{aligned}
        \mathbf{H} &= \frac{2}{N}\sum_{i=1}^N w_i \tilde{\mathbf{M}}_i^\top \tilde{\mathbf{M}}_i + 2\lambda_\mathbf{c} \mathbf{M}_c^\top \mathbf{M}_c \\
        \mathbf{g} &= \frac{2}{N}\sum_{i=1}^N w_i \tilde{\mathbf{M}}_i^\top \tilde{\mathbf{d}}_i + 2\lambda_\mathbf{c} \mathbf{M}_c^\top \mathbf{d}_c
    \end{aligned}
\end{equation}

\section{ADMM Formulation}
\label{appx:admm_formulation}

% Our framework adopts an ADMM-based solver inspired by the ReLU-QP architecture~\cite{Bishop2023ReLUQPAG}. We introduce splitting variables $\bar{\boldsymbol{\omega}}$ and $\mathbf{z}$ to decouple the quadratic objective from the non-convex scaled-rotation set $\mathcal{M} = \mathbb{R}_{++} \times \mathrm{SO}(3)$.
Our framework adopts an ADMM-based solver inspired by the ReLU-QP architecture~\cite{Bishop2023ReLUQPAG}. The physical scaled-rotation domain is the manifold
\[
\mathcal M
=
\{\alpha\mathbf Q
\mid
\alpha>0,\;
\mathbf Q\in\mathrm{SO}(3)\},
\]
where $\alpha=s^{-1}$ denotes inverse metric scale. For numerical projection, ADMM uses its closure
\[
\overline{\mathcal M}
=
\{\alpha\mathbf Q
\mid
\alpha\geq0,\;
\mathbf Q\in\mathrm{SO}(3)\}.
\]
We introduce splitting variables $\bar{\boldsymbol{\omega}}$ and $\mathbf z$ to decouple the quadratic objective from this scaled-rotation constraint.

\subsection{Problem Reformulation}
The optimization problem is reformulated as follows:
\begin{equation}
    \begin{aligned}
        \min_{\boldsymbol{\omega}, \bar{\boldsymbol{\omega}}, \mathbf{z}} \quad & \frac{1}{2}\bar{\boldsymbol{\omega}}^\top \mathbf{H} \bar{\boldsymbol{\omega}} + \mathbf{g}^\top \bar{\boldsymbol{\omega}} + \mathbb{I}_{\overline{\mathcal{M}}}(\mathbf{z}) \\
        \text{s.t.} \quad & \bar{\boldsymbol{\omega}} = \mathbf{z}, \quad \boldsymbol{\omega} = \bar{\boldsymbol{\omega}}
    \end{aligned}
\end{equation}
% where the indicator function $\mathbb{I}_{\mathcal{M}}(\mathbf{z})$ is zero if $\mathrm{mat}(\mathbf{z}) \in \mathcal{M}$ and $+\infty$ otherwise.
where the indicator function $\mathbb{I}_{\overline{\mathcal{M}}}(\mathbf{z})$ is zero if $\mathrm{mat}(\mathbf{z})\in\overline{\mathcal{M}}$ and $+\infty$ otherwise.

\subsection{Augmented Lagrangian}
The augmented Lagrangian for this system is defined as:
\begin{equation}
    \begin{aligned}
        L_{\rho, \sigma} &= \frac{1}{2}\bar{\boldsymbol{\omega}}^\top \mathbf{H} \bar{\boldsymbol{\omega}} + \mathbf{g}^\top \bar{\boldsymbol{\omega}} + \mathbb{I}_{\overline{\mathcal{M}}}(\mathbf{z}) \\
        &\quad + \frac{\sigma}{2}\|\bar{\boldsymbol{\omega}}-\boldsymbol{\omega}\|_2^2 + \boldsymbol{\mu}^\top (\bar{\boldsymbol{\omega}}-\boldsymbol{\omega}) \\
        &\quad + \frac{\rho}{2}\|\bar{\boldsymbol{\omega}}-\mathbf{z}\|_2^2 + \boldsymbol{\lambda}^\top (\bar{\boldsymbol{\omega}}-\mathbf{z})
    \end{aligned}
\end{equation}
% where $\sigma, \rho \in \mathbb{R}_{++}$ are penalty weights.
where $\sigma,\rho\in\mathbb{R}_{++}$ are the consensus and scaled-rotation penalty parameters, respectively.

\subsection{Iterative Updates}

We solve the quadratic objective using a proximal ADMM iteration with adaptive
scaled-rotation penalty $\rho_t$. Let
$\boldsymbol{\omega}\in\mathbb{R}^{9}$ denote the vectorized scaled rotation,
$\mathbf z\in\mathbb{R}^{9}$ its scaled-rotation-constrained copy, and
$\boldsymbol{\lambda}\in\mathbb{R}^{9}$ the unscaled dual variable. We fix
$\sigma=10^{-5}$ and initialize
\begin{equation}
\rho_0
=
\max\left(
\rho_{\mathrm{base}}
\operatorname{median}_{i}(H_{ii}),
\;\rho_{\min}
\right),
\end{equation}
with $\rho_{\mathrm{base}}=3\times10^{-2}$ and
$\rho_{\min}=10^{-5}$.

At iteration $t$, we use
\begin{equation}
\widetilde{\mathbf{H}}_t
=
\mathbf H + (\sigma+\rho_t)\mathbf I,
\end{equation}
and, starting from
$\boldsymbol{\omega}^{0}=\mathbf z^{0}
=\boldsymbol{\lambda}^{0}=\mathbf 0$, perform
\begin{equation}
\begin{aligned}
\boldsymbol{\omega}^{t+1}
&=
\widetilde{\mathbf{H}}_t^{-1}
\left(
-\mathbf g
+\sigma\boldsymbol{\omega}^{t}
+\rho_t\mathbf z^{t}
-\boldsymbol{\lambda}^{t}
\right), \\
\mathbf z^{t+1}
&=
\mathcal{P}_{\overline{\mathcal M}}
\left(
\boldsymbol{\omega}^{t+1}
+\rho_t^{-1}\boldsymbol{\lambda}^{t}
\right), \\
\boldsymbol{\lambda}^{t+1}
&=
\boldsymbol{\lambda}^{t}
+
\rho_t
\left(
\boldsymbol{\omega}^{t+1}-\mathbf z^{t+1}
\right).
\end{aligned}
\label{eq:asm-admm}
\end{equation}

The inverse is never formed explicitly. Instead, we solve the linear system
using a batched Cholesky factorization, which is reused while $\rho_t$ remains
unchanged and recomputed after each penalty update. Following
Eq.~\eqref{eq:rho_adaptation}, $\rho_t$ is adapted every three iterations
starting at $t=3$. We use 20 ADMM iterations at inference and return the
projected iterate $\mathbf z^t$ with the lowest original quadratic objective.

\subsection{Scaled-Rotation Projection}
% Let
% \begin{equation}
% \mathcal{M}
% =
% \left\{
% s\mathbf{R}
% \;\middle|\;
% s > 0,\;
% \mathbf{R}\in\mathrm{SO}(3)
% \right\}
% \subset\mathbb{R}^{3\times 3}.
% \end{equation}
% Since the ADMM variables are vectorized while the constraint is
% matrix-valued, for $\mathbf{u}\in\mathbb{R}^{9}$ we define
% \begin{equation}
%     \mathcal{P}_{\mathcal{M}}(\mathbf{u})
%     =
%     \operatorname{vec}
%     \left(
%         \Pi_{\mathcal{M}}
%         \left(
%             \operatorname{mat}(\mathbf{u})
%         \right)
%     \right).
% \end{equation}
The physical pose model uses the positive inverse scale $\alpha=s^{-1}>0$. For numerical projection, ADMM uses the closure
\begin{equation}
    \overline{\mathcal M}
    =
    \{\alpha\mathbf R
    \mid
    \alpha\geq0,\;
    \mathbf R\in\mathrm{SO}(3)\}
    \subset\mathbb R^{3\times3}.
\end{equation}
Since the ADMM variables are vectorized while the constraint is matrix-valued, for $\mathbf u\in\mathbb R^9$ we define
\begin{equation}
    \mathcal P_{\overline{\mathcal M}}(\mathbf u)
    =
    \operatorname{vec}
    \left(
    \Pi_{\overline{\mathcal M}}
    (\operatorname{mat}(\mathbf u))
    \right).
\end{equation}
For $\mathbf{U}=\operatorname{mat}(\mathbf{u})$, we first recover the closest
proper rotation:
\begin{equation}
    \widehat{\mathbf{R}}
    =
    \underset{\mathbf{R}\in\mathrm{SO}(3)}
    {\operatorname{argmax}}
    \left\langle
        \mathbf{R},
        \mathbf{U}
    \right\rangle_F.
\label{eq:rotation-procrustes}
\end{equation}
We compute this rotation using Horn's quaternion formulation. Define
\begin{equation}
    \mathbf{G}(\mathbf{U})
    =
    \begin{bmatrix}
        \operatorname{tr}(\mathbf{U})
        &
        \boldsymbol{\nu}(\mathbf{U})^{\top}
        \\
        \boldsymbol{\nu}(\mathbf{U})
        &
        \mathbf{U}
        +
        \mathbf{U}^{\top}
        -
        \operatorname{tr}(\mathbf{U})\mathbf{I}
    \end{bmatrix},
\end{equation}
where
\begin{equation}
    \boldsymbol{\nu}(\mathbf{U})
    =
    \begin{bmatrix}
        U_{23}-U_{32} \\
        U_{31}-U_{13} \\
        U_{12}-U_{21}
    \end{bmatrix}.
\end{equation}
The dominant unit eigenvector of the symmetric matrix
$\mathbf{G}(\mathbf{U})$ defines a quaternion, which is converted to
$\widehat{\mathbf{R}}\in\mathrm{SO}(3)$. This is equivalent to the standard
proper-rotation Procrustes solution and requires only a symmetric
$4\times4$ eigendecomposition. The exact Euclidean projection onto $\overline{\mathcal{M}}$ is
% \begin{equation}
%     \Pi_{\mathcal{M}}(\mathbf{U})
%     =
%     s^\star\widehat{\mathbf{R}},
%     \qquad
%     s^\star
%     =
%     \max
%     \left(
%         \varepsilon,
%         \frac{
%             \left\langle
%                 \widehat{\mathbf{R}},
%                 \mathbf{U}
%             \right\rangle_F
%         }{3}
%     \right).
% \label{eq:metric-projection}
% \end{equation}
\begin{equation}
    \Pi_{\overline{\mathcal M}}(\mathbf U)
    =
    \alpha^\star\widehat{\mathbf R},
    \qquad
    \alpha^\star
    =
    \max\left(
    0,\,
    \frac{
    \langle\widehat{\mathbf R},\mathbf U\rangle_F
    }{3}
    \right).
\label{eq:metric-projection}
\end{equation}
Thus, the projection jointly recovers the closest proper rotation and its optimal nonnegative scale. It is the exact metric projection used in each ADMM iteration rather than a heuristic normalization or retraction.

\section{SDP Global Optimality Certification}
\label{appx:sdp_kkt}

To certify the global optimality of the solution $\boldsymbol{\omega}^*$, the QCQP is relaxed into a Semidefinite Program (SDP) by lifting the state to $\mathbf{W} = \mathbf{w}\mathbf{w}^\top \in \mathbb{S}_{+}^{10}$, where $\mathbf{w} = [\boldsymbol{\omega}^\top, 1]^\top$. 

\subsection{Semidefinite Relaxation}
By dropping the non-convex rank-1 constraint, the resulting SDP relaxation is formulated as:
\begin{equation}
    \begin{aligned}
        \min_{\mathbf{W}\succeq \mathbf{0}} \quad & \left\langle \mathbf{Q}, \mathbf{W} \right\rangle \\
        \text{s.t.} \quad & \langle \mathbf{A}_i, \mathbf{W} \rangle = a_i, \quad i=1,\dots,7
    \end{aligned}
\label{eq:sdp_relaxation}
\end{equation}
where $\langle \mathbf{A}, \mathbf{B} \rangle = \text{tr}(\mathbf{A}^\top \mathbf{B})$ denotes the Frobenius inner product.

\subsection{Cost and Constraint Matrices}
The homogenized cost matrix $\mathbf{Q} \in \mathbb{R}^{10 \times 10}$ is constructed from the reduced quadratic form:
\begin{equation}
    \mathbf{Q} = \begin{bmatrix}
        \frac{1}{2}\mathbf{H} & \frac{1}{2}\mathbf{g} \\
        \frac{1}{2}\mathbf{g}^\top & 0 
    \end{bmatrix}
\end{equation}
The manifold $\mathcal{M}$ is enforced by seven matrices $\mathbf{A}_i$.
The equal-norm constraints (where $a_{1,2,3}=0$) are:
\begin{equation}
    \begin{aligned}
        \mathbf{A}_1 &= \text{diag}(\mathbf{I}_3, -\mathbf{I}_3, \mathbf{0}_3, 0) \\
        \mathbf{A}_2 &= \text{diag}(\mathbf{I}_3, \mathbf{0}_3, -\mathbf{I}_3, 0) \\
        \mathbf{A}_3 &= \text{diag}(\mathbf{0}_3, \mathbf{I}_3, -\mathbf{I}_3, 0)
    \end{aligned}
\end{equation}
The orthogonality constraints (where $a_{4,5,6}=0$) use symmetric off-diagonal blocks:
\begin{equation}
    \mathbf{A}_4 = \frac{1}{2} \begin{bmatrix} \mathbf{0} & \mathbf{I}_3 & \mathbf{0} & 0 \\ \mathbf{I}_3 & \mathbf{0} & \mathbf{0} & 0 \\ \mathbf{0} & \mathbf{0} & \mathbf{0} & 0 \\ 0 & 0 & 0 & 0 \end{bmatrix}, \quad
    \mathbf{A}_5 = \frac{1}{2} \begin{bmatrix} \mathbf{0} & \mathbf{0} & \mathbf{I}_3 & 0 \\ \mathbf{0} & \mathbf{0} & \mathbf{0} & 0 \\ \mathbf{I}_3 & \mathbf{0} & \mathbf{0} & 0 \\ 0 & 0 & 0 & 0 \end{bmatrix}
\end{equation}
\begin{equation}
    \mathbf{A}_6 = \frac{1}{2} \begin{bmatrix} \mathbf{0} & \mathbf{0} & \mathbf{0} & 0 \\ \mathbf{0} & \mathbf{0} & \mathbf{I}_3 & 0 \\ \mathbf{0} & \mathbf{I}_3 & \mathbf{0} & 0 \\ 0 & 0 & 0 & 0 \end{bmatrix}, \quad
    \mathbf{A}_7 = \text{diag}(0, \dots, 1)
\end{equation}
with $a_7 = 1$ to enforce the homogeneity of the state vector.

\subsection{Estimation of Lagrange Multipliers}
From the QCQP stationarity condition, the Lagrange multiplier vector $\boldsymbol{\eta} \in \mathbb{R}^7$ must satisfy $\sum_{i=1}^7 \eta_i \mathbf{A}_i \mathbf{w}^* = \mathbf{Q} \mathbf{w}^*$. We define the matrix $\mathbf{M} = [\mathbf{A}_1 \mathbf{w}^*, \dots, \mathbf{A}_7 \mathbf{w}^*] \in \mathbb{R}^{10 \times 7}$ and solve for $\boldsymbol{\eta}$ via least squares:
\begin{equation}
    \boldsymbol{\eta} = \mathbf{M}^\dagger (\mathbf{Q} \mathbf{w}^*)
\end{equation}

\subsection{Global Optimality Certificate}
The dual slack matrix is computed as $\mathbf{S} = \mathbf{Q} - \sum_{i=1}^7 \eta_i \mathbf{A}_i$. A candidate solution $\mathbf{w}^*$ is certified as globally optimal if:
\begin{equation}
    \mathbf{S}\mathbf{w}^* = \mathbf{0} \quad \text{and} \quad \mathbf{S} \succeq \mathbf{0}
\end{equation}
Verification of the second condition is performed by ensuring $\lambda_{\min}(\mathbf{S}) \geq -\epsilon$, where $\epsilon$ is a small numerical tolerance.

\subsection{Rank-1 Recovery}
If Eq.~\eqref{eq:sdp_relaxation} is solved via an SDP solver, the vector $\mathbf{w}^*$ is recovered from the principal eigenvector $\mathbf{v}_1$ of $\mathbf{W}^*$ corresponding to the largest eigenvalue $\lambda_1$:
\begin{equation}
    \mathbf{w}^* = \frac{\sqrt{\lambda_1} \mathbf{v}_1}{(\sqrt{\lambda_1} \mathbf{v}_1)_{10}}
\end{equation}
This ensures the final solution satisfies $w_{10}^* = 1$.

\section{Training and Evaluation Details}
\label{appx:training_eval_details}

% \subsection{Data Separation and Benchmark Protocols}  
% The Omni6DPose ASM library is constructed from all $4{,}385$ standardized CAD models. SOPE train and test images share this library, so SOPE evaluates held-out observations of known CAD instances rather than unseen-instance shape generalization. ROPE contains separately captured real instances, and no ROPE images are used for training or model selection. HouseCat6D uses its official train/validation/test split, with category priors constructed exclusively from training CADs.

\subsection{SOPE Training}
\label{appx:sope_training}
SOPE~\cite{Zhang2024Omni6DPoseAB} models are trained with AdamW using a learning rate of $10^{-4}$ and a batch size of 64. Weight decay is $10^{-4}$ for task-specific modules and zero for adapters. We set $\lambda_{\mathrm{vis}}=\lambda_{\mathrm{depth}}=\lambda_{\mathrm{pose}}=0.1$ and $\lambda_{\mathrm{trans}}=1.0$. LoRA and DoRA use rank-32 adapters on the same 48 projection layers of the frozen RGB-D backbone. 
%During training, we jointly rotate the RGB-D input and corresponding labels, and apply random depth and segmentation-based dropout to improve robustness to viewpoint variation and missing observations.
During training, we apply random in-plane rotations and random crops to the RGB-D inputs. We additionally apply RGB photometric augmentation and random erasing, depth dropout and noise, and independent segmentation-based background masking of RGB and depth with perturbed masks to improve robustness to appearance variation and missing or corrupted observations.

\subsection{HouseCat6D Training and Evaluation}
\label{appx:housecat_training}
HouseCat6D~\cite{jung2024housecat6d} (HC6D) uses the official 34/2/5 train/validation/test split. Both training regimes use rank-32 LoRA, batch size 64, AdamW learning rates of $10^{-4}$ for task-specific modules and $3\times10^{-5}$ for adapters, and task-module weight decay $10^{-3}$. We use the same rotation and dropout augmentations as for SOPE. We use unit-weight correspondence loss, $\lambda_{\mathrm{vis}}=10^{-3}$, $\lambda_{\mathrm{depth}}=\lambda_{\mathrm{pose}}=0.1$, and symmetry-aware pose loss with rotation/translation weights $2.0/1.5$ and time weighting $(1-t)^2$. The HC6D model is trained without SOPE initialization, whereas SOPE$\rightarrow$HC6D initializes from the SOPE-trained LoRA model.

Evaluation follows the public AG-Pose GT-instance protocol~\cite{Lin_2024_CVPR}. Both variants use four-step distillation, $H=40$, training-CAD ASMs, and unmasked completed depth. Sensor-assisted depth refinement is disabled.

\subsection{Shape-Completion Protocol}
\label{appx:completion_protocol}
We evaluate completion on the SOPE test set at image stride 3 ($339{,}840$ instances) and on $14{,}077$ RealNOCS \texttt{real\_test} instances matched to released DR-Pose detections at mask IoU $\geq0.5$. Predictions are compared with posed mesh ground truth in the camera frame using fixed manifests, seeds, and mesh samples. Evaluation uses $1{,}152$ points on SOPE and 642 points on RealNOCS.

We report F1@1\%, CD-L1, and HD95, with the F1 threshold set to 1\% of the ground-truth bounding-box diagonal. Point-cloud baselines receive 2,048 instance-mask-segmented depth points, whereas ASM-6D receives an RGB-D ROI without mask-segmented points as model input. The comparison therefore reflects each method's native perception interface rather than an input-matched completion protocol. ``w/o prior'' denotes raw generated correspondences, while ``w/ prior'' denotes the ASM-fitted output.

\subsection{Tracking Protocol}
\label{appx:tracking_protocol}
For tracking, the first frame estimates pose, metric scale, and shape from four correspondence hypotheses generated with two denoising steps. The estimated scale and shape are retained over the sequence. Each subsequent frame generates new correspondence hypotheses and selects the pose using geometric fit and motion consistency. Translation and rotation are smoothed using One-Euro filters~\cite{casiez20121}.

The HO3D evaluation uses sequences \texttt{ABF10}, \texttt{BB10}, \texttt{GPMF10}, \texttt{MC1}, \texttt{SM2}, \texttt{SMu1}, and \texttt{SS1}, comprising seven YCB objects and 448 scored frames. Both methods receive the same RGB-D frames, camera intrinsics, and oracle object silhouette; neither receives ground-truth pose. ASM-6D uses the corresponding category-level ASM, whereas FoundationPose receives the exact instance CAD.

The high-motion evaluation uses a dynamic mug sequence with OptiTrack ground truth, with mean translation and rotation speeds of $0.197\,\mathrm{m\,s^{-1}}$ and $88.2^\circ\,\mathrm{s^{-1}}$, respectively. Both methods receive the same RGB-D frames, intrinsics, and offline SAM2.1~\cite{ravi2025sam} masks. FoundationPose uses the exact YCB \texttt{025\_mug} model. We evaluate native tracking and periodic re-registration every 30, 60, and 120 frames.

\section{Model and Inference Ablations}
\label{app:additional_ablations}

\begin{table}[t]
\caption{Ablation study of ASM-6D.}
\label{tab:combined_ablation}
\centering
\footnotesize
\setlength{\tabcolsep}{1.2pt}
\renewcommand{\arraystretch}{1.03}

\begin{tabular}{l|ccc|cccc}
\toprule
\toprule
\multirow{2}{*}{Configuration}
& \multicolumn{3}{c|}{AUC $\uparrow$}
& \multicolumn{4}{c}{VUS $\uparrow$} \\
& IoU$_{25}$ & IoU$_{50}$ & IoU$_{75}$
& $5^\circ$/2cm & $5^\circ$/5cm
& $10^\circ$/2cm & $10^\circ$/5cm \\
\hline

\multicolumn{8}{l}{\textit{\textbf{A)} Regression vs.\ flow matching, frozen $H = 1$}} \\

Regression
& 27.87 & 11.58 & 0.93
& 4.38 & 6.81 & 10.88 & 16.98 \\

Flow Matching
& \textbf{43.91} & \textbf{25.91} & \textbf{5.10}
& \textbf{10.29} & \textbf{14.45} & \textbf{20.56} & \textbf{29.24} \\

\hline
\multicolumn{8}{l}{\textit{\textbf{B)} Hypothesis count, distilled model}} \\

$H=1$
& 47.94 & 30.07 & 7.11
& 14.67 & 20.54 & 26.70 & 37.73 \\

$H=4$
& 48.54 & 30.59 & 7.35
& 15.17 & 21.20 & 27.38 & 38.65 \\

$H=8$
& 48.70 & 30.73 & 7.42
& 15.30 & 21.37 & 27.55 & 38.86 \\

$H=16$
& 48.83 & 30.84 & 7.48
& 15.38 & 21.49 & 27.66 & 39.01 \\

$H=32$
& 48.93 & 30.92 & 7.51
& 15.44 & 21.56 & 27.74 & 39.11 \\

$H=40$
& \textbf{48.96} & \textbf{30.95} & \textbf{7.53}
& \textbf{15.46} & \textbf{21.58}
& \textbf{27.76} & \textbf{39.14} \\

\hline
\multicolumn{8}{l}{\textit{\textbf{C)} ADMM iterations, distilled model}} \\

3 iterations
& 46.53 & 28.27 & 6.18
& 13.30 & 18.74 & 25.43 & 36.29 \\

5 iterations
& 48.28 & 30.09 & 6.95
& 14.65 & 20.48 & 26.98 & 38.18 \\

10 iterations
& 48.87 & 30.83 & 7.40
& 15.35 & 21.42 & 27.68 & 39.02 \\

20 iterations
& \textbf{48.96} & \textbf{30.95} & \textbf{7.53}
& \textbf{15.46} & \textbf{21.58}
& \textbf{27.76} & \textbf{39.14} \\

\hline
\multicolumn{8}{l}{\textit{\textbf{D)} Global-depth refinement, distilled model}} \\

0 updates
& 48.61 & 30.64 & 7.31
& 15.26 & 21.51 & 27.38 & 39.00 \\

1 update
& \textbf{48.96} & \textbf{30.95} & \textbf{7.53}
& \textbf{15.46} & \textbf{21.58}
& \textbf{27.76} & \textbf{39.14} \\

2 updates
& 48.94 & 30.89 & 7.52
& 15.45 & 21.56 & \textbf{27.76} & 39.11 \\

3 updates
& 48.88 & 30.81 & 7.48
& 15.41 & 21.54 & 27.68 & 39.07 \\

\hline
\multicolumn{8}{l}{\textit{\textbf{E)} Hypothesis selection, distilled model}} \\

ASM
& \textbf{48.96} & \textbf{30.95} & \textbf{7.53}
& \textbf{15.46} & \textbf{21.58}
& \textbf{27.76} & 39.14 \\

ASM + 0.25 Depth
& 48.94 & 30.93 & 7.52
& 15.45 & \textbf{21.58}
& 27.75 & \textbf{39.15} \\

ASM + 0.5 Depth
& 48.92 & 30.91 & 7.51
& 15.43 & 21.57 & 27.72 & 39.13 \\

ASM + 1.0 Depth
& 48.89 & 30.87 & 7.50
& 15.40 & 21.52 & 27.68 & 39.08 \\

% Remove oracle?
% & -- & -- & --
% & -- & -- & -- & -- \\

% Oracle best-of-$40$
% & 49.69 & 32.10 & 8.58
% & 22.91 & 31.73 & 34.61 & 48.25 \\

\hline
\multicolumn{8}{l}{\textit{\textbf{F)} Shape prior and PCA rank}} \\

Icosphere
& 16.99 & 5.50 & 0.64
& 4.88 & 8.98 & 10.94 & 21.33 \\

Category mean
& 39.35 & 19.82 & 2.76
& 11.76 & 17.68 & 22.70 & 34.63 \\

ASM, $K \leq 4$
& 47.57 & 29.12 & 6.64
& 14.93 & 21.03 & 27.02 & 38.51 \\

ASM, $K \leq 8$
& 48.44 & 30.28 & 7.24
& 15.34 & 21.45 & 27.55 & 38.94 \\

ASM, $K \leq 16$
& 48.82 & 30.78 & 7.50
& 15.43 & 21.54 & 27.69 & 39.07 \\

ASM, $K \leq 48$
& \textbf{48.96} & \textbf{30.95} & \textbf{7.53}
& \textbf{15.46} & \textbf{21.58}
& \textbf{27.76} & \textbf{39.14} \\
\bottomrule
\bottomrule
\end{tabular}

\vspace{0.5mm}
\raggedright
\scriptsize
AUC and VUS are percentages. Bold indicates the best result within each ablation.
\end{table}

Table~\ref{tab:combined_ablation} evaluates each component by varying one factor at a time. Unless stated otherwise, we use the four-step distilled model, $H=40$, one global-depth refinement update, the visibility-weighted mean ASM selector, a category-specific ASM, and 20 ADMM iterations. The solver returns the feasible iterate with the lowest reduced objective. All comparisons use identical data ordering and random seeds.

\subsection{Regression Versus Flow Matching}
We compare deterministic regression and flow matching at $H=1$ to exclude the effect of sampling multiple correspondence sets. Flow matching increases IoU$_{50}$ AUC from $11.58$ to $25.91$ and $10^\circ$/5\,cm VUS from $16.98$ to $29.24$, demonstrating that the improvement is not attributable solely to multi-hypothesis inference.

\subsection{Hypothesis Budget}
We evaluate $H\in\{1,4,8,16,32,40\}$. Accuracy improves monotonically but saturates at moderate $H$. Reducing the budget from $40$ to $4$ decreases $10^\circ$/5\,cm VUS by $0.49$ points. In return, known-class throughput increases from $10.26$ to $30.24$ FPS, while exhaustive 156-prior throughput increases from $6.82$ to $20.65$ FPS. We use $H=40$ for maximum accuracy and report smaller budgets as an accuracy--throughput tradeoff.

\subsection{ADMM Iteration Budget}
We evaluate 3, 5, 10, and 20 ADMM iterations using an identical correspondence set, global-depth refinement, and selection. Most gains occur within 10 iterations. Increasing the budget from 10 to 20 improves $10^\circ$/5\,cm VUS by only $0.12$ points. At $H=40$, known-class throughput changes from $10.33$ to $10.26$ FPS, and exhaustive-prior throughput changes from $6.98$ to $6.82$ FPS. We therefore use 20 iterations as a conservative accuracy-oriented default.

\subsection{Global-Depth Refinement}
A single sensor-assisted refinement update provides the best overall result, increasing IoU$_{50}$ AUC from $30.64$ to $30.95$ and
$10^\circ$/5\,cm VUS from $39.00$ to $39.14$ relative to no refinement. Further updates provide no consistent improvement and slightly reduce several metrics. We therefore apply one update in the standard configuration.

\subsection{Hypothesis Selection Using Depth}
As an ablation, we additionally compare the fitted point depth with the
observed depth image. Let
$(\hat{u}_i^{h,j},\hat v_i^{h,j},\hat d_i^{h,j})$ denote the projection and depth of fitted
point $\hat {\mathbf y}_i^{h,j}$, and define
\begin{equation}
\begin{aligned}
    d_{i}^{h,j}
    &=
    \mathcal D(\hat u_i^{h,j},\hat v_i^{h,j}),
    \\
    e_{D,i}^{h,j}
    &=
    \frac{
        \left|
            \hat d_i^{h,j}-d_{i}^{h,j}
        \right|
    }{
        d_{i}^{h,j}
    }
    \begin{cases}
        \beta_{\mathrm{occ}},
        &
        \hat d_i^{h,j}-d_{i}^{h,j}
        >
        \epsilon_{\mathrm{occ}},
        \\
        1,
        & \text{otherwise}.
    \end{cases}
\end{aligned}
\label{eq:selection_depth_residual}
\end{equation}
The factor $\beta_{\mathrm{occ}}<1$ reduces the penalty for fitted points
behind the observed surface. Over valid depth queries $\mathcal I_{h,j}$,
\begin{equation}
    \gamma_D(h,j)
    =
    \frac{
        \sum_{i\in\mathcal I_{h,j}}
        w_i^h e_{D,i}^{h,j}
    }{
        \sum_{i\in\mathcal I_{h,j}}
        w_i^h
    }.
\label{eq:mean_depth_score}
\end{equation}
Let $\mathcal R(\cdot)$ denote normalized rank among valid candidates. The
diagnostic fusion score is
\begin{equation}
    S_{\lambda_D}(h,j)
    =
    \mathcal R\!\left(\gamma_{\mathrm{ASM}}(h,j)\right)
    +
    \lambda_D
    \mathcal R\!\left(\gamma_D(h,j)\right).
\label{eq:depth_fusion_score}
\end{equation}
We evaluate $\lambda_D\in\{0.25,0.5,1.0\}$ only as an ablation. Adding fitted-depth consistency provides no consistent benefit: a weight of $0.25$ changes $10^\circ$/5\,cm VUS by only $+0.01$ points while slightly reducing IoU AUC, and larger weights degrade performance. We therefore retain the mean-ASM residual selector in Eq.~\eqref{eq:deployed_selection}.

\begin{figure}[t]
    \centering
    \includegraphics[width=\columnwidth]
    {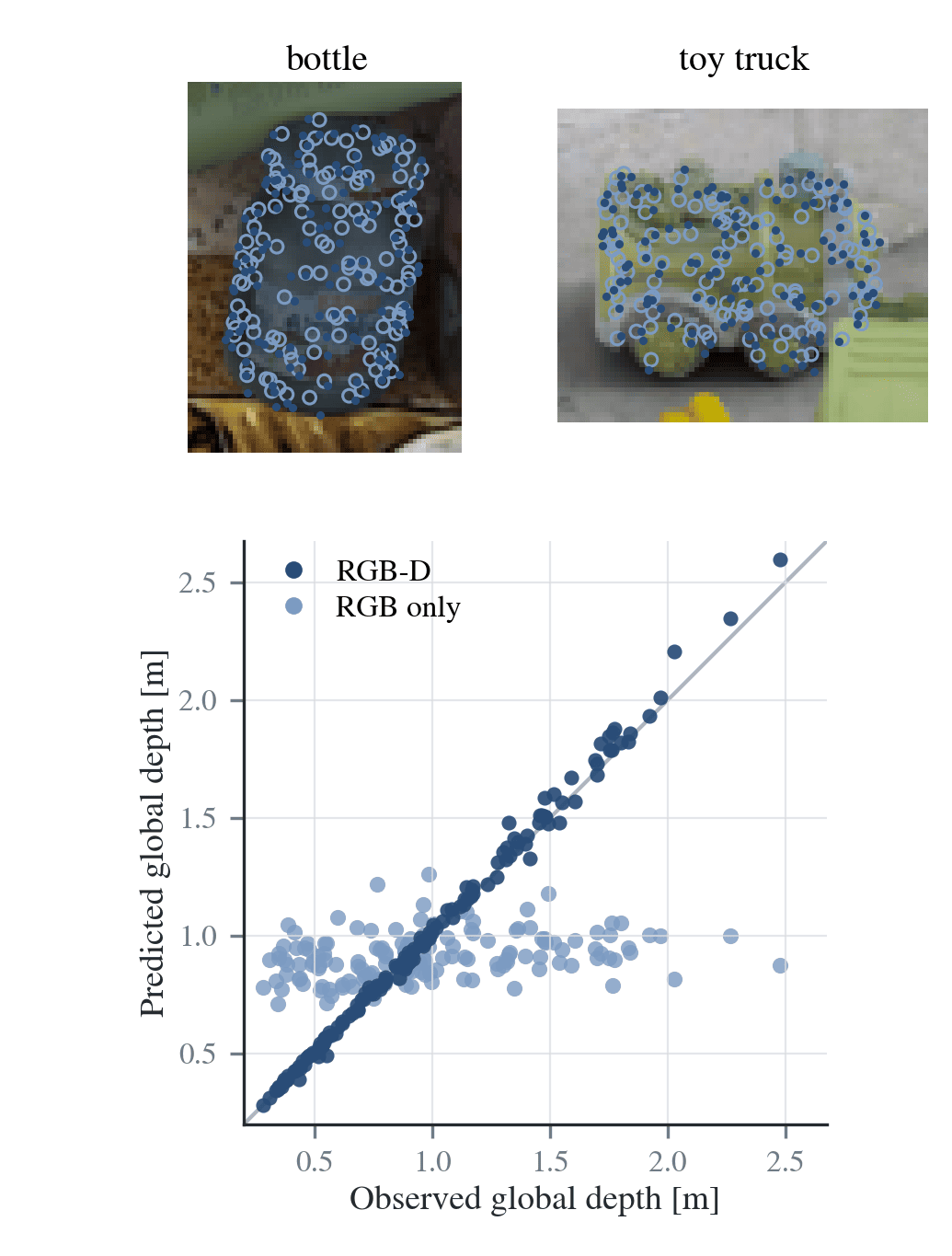}
    \caption{RGB-only input ablation. Top: representative paired RGB-D and RGB-only predictions. Bottom: predicted global depth across $50$ objects compared with observed metric depth.}
    \label{fig:rgb-only-ablation}
\end{figure}

\subsection{Role of the Shape Prior and Active Variation}
We decompose the contribution of the shape prior into category-specific mean geometry and intra-category PCA deformation. First, we replace each category-specific ASM with the same indexed 642-vertex icosphere. This preserves the model outputs, persistent correspondence indices, sampled correspondence sets, global-depth refinement, fitting procedure, ADMM configuration, and visibility-weighted mean fitting-residual selector, while removing both the category-specific mean and deformation modes.

We then vary the PCA rank while retaining the category-specific mean. For a rank cap $K$, category $j$ uses its first $\min\!\left(K, K_j^{\mathrm{avail}}\right)$ available modes, with a maximum deployed rank of 48. Setting $K=0$ therefore fits only pose and scale to the category mean, without optimizing instance-specific deformation. All rank variants are evaluated using exactly the same sampled correspondence sets and inference settings.

The category mean substantially improves over the generic icosphere. IoU$_{50}$ AUC increases from $5.50$ to $19.82$, while ($10^\circ$/5\,cm) VUS increases from $21.33$ to $34.63$. This establishes the importance of class-specific mean geometry beyond persistent correspondence indices alone. The full ASM ($K \leq 48$) further improves IoU$_{50}$ AUC from $19.82$ to $30.95$ and ($10^\circ$/5\,cm) VUS from $34.63$ to $39.14$, demonstrating the additional benefit of optimizing intra-category PCA deformation. The first four modes recover most of this improvement, reaching $29.12$ IoU$_{50}$ AUC and $38.51$ ($10^\circ$/5\,cm) VUS, while performance largely saturates by 16 modes.

\subsection{RGB-Only Input Ablation}
\label{appx:rgb-only-ablation}

We compare paired RGB-D and RGB-only inference on $50$ SOPE test objects from
$41$ categories. Each pair uses the same crop, checkpoint, Gaussian initializations, $H=8$ correspondence sets, and four denoising steps. For RGB-only inference, depth tokens are removed using the encoder's native masking mechanism. Ground-truth masks define the crops but are not provided to the model.

RGB alone preserves much of the correspondence structure. Relative to RGB-D, the median image-plane displacement is $3.34$ pixels, the median discrepancy after 3D similarity alignment is $11.0\%$ of the RGB-D shape radius, and the median mean absolute difference in relative log-depth is $0.0084$. This suggests that RGB provides a strong prior over canonical layout and relative shape.

Depth is instead crucial for metric calibration and observation-dependent geometry. The median global-depth error increases from $2.48\%$ to $28.8\%$ without depth, while its correlation with observed depth drops from $0.995$ to $0.194$. Fig.~\ref{fig:rgb-only-ablation} visualizes the effect of removing the depth tokens. Median visibility also decreases from $0.370$ to $0.173$, and the normalized known-class ASM residual rises from $2.93\%$ to $6.32\%$, with degradation on $96\%$ of the objects. Thus, RGB largely determines the coarse correspondence structure, whereas depth anchors metric scale, visibility, and fine 3D consistency. Because the model is trained with RGB-D, this ablation measures modality reliance rather than the capability of a separately trained RGB-only model.

% \end{thebibliography}
% \clearpage
\bibliography{Star} % Entries are in the Star.bib file


% \end{IEEEbiography}


%\newpage



%\vfill


\end{document}